\documentclass[lettersize,journal]{IEEEtran}
\usepackage{amsmath,amssymb, amsthm, amsfonts}
\usepackage[ruled,linesnumbered]{algorithm2e}
\usepackage{algorithmic}
\usepackage{array}
\usepackage[caption=false,font=footnotesize,labelfont=rm,textfont=rm]{subfig}
\usepackage{cite}
\usepackage{enumitem}[inline]
\usepackage{textcomp}
\usepackage{color, xcolor}
\usepackage{stfloats}
\usepackage{url}
\usepackage{verbatim}
\usepackage{graphicx}
\usepackage{hyperref} % 颜色可调

\usepackage{soul}

\usepackage{amsthm}
\theoremstyle{remark}
\newtheorem{remark}{Remark}

\theoremstyle{definition}
\newtheorem{theorem}{Theorem}
\newtheorem{lemma}{Lemma}
\newtheorem{assumption}{Assumption}

\definecolor{mygray}{RGB}{200,200,200}
\sethlcolor{mygray}

\usepackage[T1]{fontenc}
\usepackage{booktabs}

\usepackage{multirow}

\def\BibTeX{{\rm B\kern-.05em{\sc i\kern-.025em b}\kern-.08em
    T\kern-.1667em\lower.7ex\hbox{E}\kern-.125emX}}
\usepackage{balance}
\begin{document}
\title{
Faster Convergence on Heterogeneous Federated Edge Learning: An Adaptive Clustered Data Sharing Approach
}
\author{\IEEEauthorblockN{Gang Hu, Yinglei Teng, \textit{Senior Member, IEEE}, Nan Wang, and Zhu Han, \textit{Fellow, IEEE}}
\thanks{This work was supported in part by the National Natural Science Foundation
of China under Grant No. 62171062. Part of this work has been presented
in IEEE International Conference on Communications (ICC) \cite{Clustered_data_sharing_ICC}.

Gang Hu, Nan Wang, and Yinglei Teng are with the Beijing Key Laboratory of Work Safety Intelligent Monitoring, Beijing University of Posts and Telecommunications (BUPT), Xitucheng Road No.10, Beijing, China, 100876. (Email: hugang@bupt.edu.cn; wangnan\_26@bupt.edu.cn; lilytengtt@gmail.com).

Zhu Han is with the Department of Electrical and Computer Engineering at the University of Houston, Houston, TX 77004 USA. (e-mails: hanzhu22@gmail.com)
}}

%\markboth{Journal of \LaTeX\ Class Files,~Vol.~18, No.~9, April~2023}%
%{How to Use the IEEEtran \LaTeX \ Templates}

\maketitle
\begin{abstract}
Federated Edge Learning (FEL) emerges as a pioneering distributed machine learning paradigm for the 6G Hyper-Connectivity, harnessing data from the IoT devices while upholding data privacy. However, current FEL algorithms struggle with non-independent and non-identically distributed (non-IID) data, leading to elevated communication costs and compromised model accuracy. To address these statistical imbalances, we introduce a clustered data sharing framework, mitigating data heterogeneity by selectively sharing partial data from cluster heads to trusted associates through sidelink-aided multicasting. The collective communication pattern is integral to FEL training, where both cluster formation and the efficiency of communication and computation impact training latency and accuracy simultaneously. To tackle the strictly coupled data sharing and resource optimization, we decompose the optimization problem into the clients clustering and effective data sharing subproblems. Specifically, a distribution-based adaptive clustering algorithm (DACA) is devised basing on three deductive cluster forming conditions, which ensures the maximum sharing yield. Meanwhile, we design a stochastic optimization based joint computed frequency and shared data volume optimization (JFVO) algorithm, determining the optimal resource allocation with an uncertain objective function. The experiments show that the proposed framework facilitates FEL on non-IID datasets with faster convergence rate and higher model accuracy in a resource-limited environment. 
\end{abstract}


\section{Introduction}
\makeatletter
\newcommand{\rmnum}[1]{\romannumeral #1}
\newcommand{\Rmnum}[1]{\expandafter\@slowromancap\romannumeral #1@}
\makeatother

In modern wireless networks, the proliferation of advanced sensors has engendered a significant surge in data generation, spanning diverse modalities such as texts, images, videos, and audio streams \cite{Survey_FL_in_MEC}. This influx of big data is conventionally offloaded and stored on cloud platforms, often coupled with Artificial intelligence (AI) techniques to facilitate a series of intelligent services. However, the privacy
leakage problem becomes critical when considering the exposure of sensitive raw data to potential attacks and cyber threats, especially after certain legislations on data privacy are introduced, e.g., the General Data Protection Regulation (GDPR) \cite{GDPR}. This predicament causes data owners to become increasingly privacy-sensitive and hesitant to share their data. As a solution, a novel distributed learning paradigm, termed Federated Learning (FL), has been introduced to construct a high-quality model by leveraging locally-trained models instead of directly sharing raw local data \cite{FedAvg}.

%All devices in the FL system collectively reap the benefits of shared models trained from the rich data, without the need to centrally store it \cite{3}. In particular, all devices iteratively train a local DL model using their private local data and forward their models to a central server. The server then aggregates them into a global model, and sends it back to devices for the subsequent round of training. In this way, the FL architecture allows devices to only share their trained models thus avoid revealing their data.

%In the framework of FL, a number of federated rounds are required between clients and edge server to coordinately achieve a target accuracy [ref]. Due to the millions of parameters in the complex deep learning model, e.g., GPT-3.5 has 175 billion parameters [ref], transmitting such high-dimensional model would result in significant communication costs. However, this costs are more serious when data across the devices is non-independent and identically distributed (non-IID). Practically, due to the decentralized and geographically dispersed nature of edge devices, data within a federated learning environment displays significant variations. According to research studies \cite{4,16}, as data heterogeneity increases, the convergence of the FL algorithm slows down, and more communication rounds are required to achieve the desired accuracy. Therefore, the heterogeneous data based on collaborative training becomes the bottleneck of FL when applied in the actual IoT environment.

In the FL framework, a number of federated rounds are required between clients and servers to collaboratively achieve desired accuracy \cite{FedAvg,FL_with_noniid,Survey_Advance_and_Open_Problem_in_FL}. Given the nature of deep learning models containing millions of parameters, e.g., the Large Language Model (LLM) GPT-3.5 has 175 billion parameters \cite{Large_AI_Model}, transmitting such high-dimensional models results in a substantial communication costs.
By balancing communication and computation resources of edge devices, Federated Edge learning (FEL) has been proposed to reduce latency and energy consumption in wireless networks \cite{FEEL}.
Yet the training costs are still serious especially when training with non-independent and non-identically distributed (non-IID) data across devices in FEL.
In practice, due to the decentralized and geographically dispersed nature of edge devices, data within a federated learning environment displays significant variations and samples imbalance, i.e., statistical heterogeneity. Extensive research studies \cite{FL_on_noniid_Reinforce_learning,Convergence_of_FedAvg_noniid,FL_with_noniid_in_WN,Survey_FL_noniid_Experimental_study} have shown that as the degree of heterogeneity increases, the convergence of the FEL algorithm slows down, and more communication rounds are required for the desired accuracy. Therefore, the time consumption and weak accuracy caused by the non-IID data become the bottleneck of FEL applied in real-world IoT applications.

There is plenty of research trying to solve the statistical heterogeneity problem, which can be categorized into two types: 1) model-driven and 2) data-driven \cite{ye2023heterogeneousFL}.
The \emph{model-driven} methods refer to designing the internal structure and parameters of the models to mitigate performance loss. Li \emph{et al}. \cite{FedProx} adds a proximal term to the local loss function to constrain the distance between the local model and the global model. \cite{FedNova} and \cite{Scaffold} correct the update drift in local training and provide convergence guarantees for non-IID FL. Instead of training a single global model with well generalization, \cite{Clustered_FL,Clustered_FL_Sattler,Fedcluster} aim at grouping the client into clusters and training personalized models based on data distribution. These works assume that there is intra-cluster similarity and inter-cluster dissimilarity in clients’ data distributions. \cite{Data_free_KD_for_heteroFL} and \cite{FedMD} provide the different model structures for heterogeneous clients by knowledge distillation techniques.
\emph{Data-driven} approaches manipulate and augment the local data to make distribution more homogenous. Zhao \emph{et al.} \cite{FL_with_noniid,FL_with_noniid_in_WN,D2D_data_sharing_DL_WN}, develop a method that shares a small subset of local data with all devices to improve model accuracy. \cite{Hybrid_FL_for_WN} and \cite{FedAux} propose a hybrid learning mechanism wherein the server collaboratively trains the model with an approximately IID dataset uploaded from the clients. \cite{Data_free_KD_for_heteroFL} employs GAN technology to ensemble data information in a data-free manner, broadcasting the generator to all users and regulating local updates to mitigate bias.
Both approaches entail sacrificing additional communication or computational resources in order to achieve accuracy improvements.

To mitigate the communication cost, some other studies try to design resource-efficient FEL system which minimize FL loss while optimizing resource efficiency. Chen \emph{et al.}{\cite{chen2020joint}} introduce a joint learning and communication framework considering resource allocation and user selection.{\protect~\cite{MultiAgentFL}} proposes a multi-agent reinforcement learning-based framework for optimizing device selection and resource allocation in distributed industrial IoT networks.{\protect~\cite{SAA}} proposes a Cluster-based Parallel Split Learning (CPSL) scheme to reduce training latency through a combination of parallel intra-cluster and sequential inter-cluster learning. These approaches These methods optimize communication resources in FEL but not considering the impact of distrusted data distribution on training performance.{\protect~\cite{FL_on_noniid_Reinforce_learning}} analyzes the implicit connection between data distribution and FEL training, and proposes a reinforcement learning-based mechanism for user selection. 
Zhao \emph{et al.}{\protect~\cite{FL_with_noniid_in_WN}} jointly minimizes the accuracy loss, delay and energy of FEL with data sharing to mitigate the impact of non-IID data.{\protect~\cite{FL_noniid_Auction_Approach}} proposes an auction approach to minimize the training and sharing cost in non-IID FEL. These frameworks rely on uploading local data to the server, which raises concerns about privacy leakage.

%Unlike the previous works, we propose a framework which reduces the degree of non-IID data and does not rely on a central server or even additional datasets. This framework solves the non-IID challenge by selecting some devices as cluster heads to share partial data with trustable associates. In non-IID FL, devices with high-quality data can accelerate FL training. Employing them as cluster heads, the data heterogeneity can be largely shrunk by data sharing. Moreover, the communication overhead and privacy issues can be well regulated by sparing data with trustworthy and reliable partners in the Device-to-Device (D2D) multicast technology.

{Although model-driven, data-driven, and system-level design approaches have attempted to address the statistical heterogeneity issue, there remain significant technical limitations deserving further research.} \emph{Firstly}, model-driven methods demand additional computing resources and show unclear improvements relying on specific  hyperparameters. The increased complexity of local training adds to the difficulty of applying FL in resource-constrained wireless networks. \emph{Secondly,} much work \cite{FL_noniid_Auction_Approach,FL_with_noniid,FL_with_noniid_in_WN,Hybrid_FL_for_WN} demonstrate that data sharing is a straightforward and efficient method for directly mitigating model quality degradation. However, most methods work under the strongly assumption of a publicly available proxy data source on the server side \cite{FedMD,FL_noniid_Auction_Approach}. Building a pre-processed and labeled public dataset is a time-consuming and expensive task.  \emph{Lastly}, the communication costs and privacy concerns of data sharing cannot be ignored. Exchanging data unavoidably raises information leakage, additional latency, and energy consumption. {To address these limitations, we adopt direct data exchange among trusted users using high-speed mmWave multicasting technology. This removes the dependency on
pre-prepared datasets or hyperparameters, reducing both privacy risks and communication overhead.}

Based on the above analysis, this work proposes a data sharing framework that allows partial data exchanged among trusted user groups to reduce the degree of data heterogeneity. In the real-world scenario, there are certain devices with high-quality data that can accelerate FEL training. By the aid of the high-bandwidth mmWave multicasting technology \cite{Muticasting_Survey}, these nodes can share the private data with trustable neighbors in close proximity communication (i.e., sidelink) directly without going through the base station (BS). These sidelink transmissions boost network ulity while saving spectrum resource since it enables the simultaneous delivery of content to multiple users through a single transmission \cite{multicasting}.
By selectively exchanging data with trustworthy and dependable partners via sidelink-aided technology, communication overhead and privacy concerns can be effectively regulated. Moreover, the goals of this paper are not only to mitigate the statistical heterogeneity issue, but also to accelerate FEL training by exploiting data and computility of edge network.
The key contributions of this paper are summarized as follows:

\begin{itemize}
\item {\textbf{Clustered data sharing FEL framework for eliminating statistical heterogeneity:} {We propose a clustered data sharing framework which can evaluate and eliminate the impacts of statistical heterogeneity in FEL. Sidelink-aided multicasting is used as an enabling technology of wireless network for FL to achieve faster training and higher accuracy. By appropriately modifying the data distribution function, the framework can be generalized to other non-IID causes, demonstrating its robustness and scalability. Moreover, this framework can also serve as a data preparation procedure applicable to model-driven non-IID FL algorithms.
    }}
\item {\textbf{Fast convergence with joint communication, computation configuration:} An optimization problem is formulated to minimize the overall delay of the FL task with clustered data sharing by jointly considering communication, computation, and privacy. Here, devices are allowed to trustfully cluster and exchange data to accelerate FEL training convergence. The shared data volume and computed frequency are jointly optimized to achieve trade-off among sharing delay, communication delay, and computation delay over the training rounds.}
\item {\textbf{An innovative low-complexity clustering algorithm utilizing constrainted graph for no closed-form problem:} By quantifying the statistical heterogeneity, the minimization delay problem is transformed into minimizing the distribution distance while adhering to privacy and communication requirements. We theoretically analyze this intractable problem from individual, intra-cluster, and inter-cluster perspectives. Subsequently, by transforming the original subproblem into a constrained graph, we devise a clustering method named distribution-based adaptive clustering algorithm (DACA) to facilitate efficient cluster formation.}
\item {\textbf{A SSCA optimization algorithm for computed frequency and shared data volume allocation problem:}
Due to the uncertain gain of data sharing on training round, the objective function with respect to shared data volume is non-closed form. Through theoretical convergence analysis, we provide a functional relationship that captures the relationship between data distribution and convergence rate. Then we employ the stochastic successive convex approximation (SSCA) algorithm to solve the nonconvex optimization problem with estimated parameters. Through solving the surrogate problem iteratively, the proposed joint computed frequency and shared data volume optimization (JFVO) algorithm is guaranteed to convergence to a stationary solution.
}
\item {\textbf{Performance evaluation from various perspectives:} Our proposed approaches are evaluated by applying classic image classification tasks to the MNIST and CIFAR-10 datasets. The clustered data sharing FL with DACA can effectively improve the accuracy in different non-IID cases and significantly reduce the total training task delay. Moreover, The experimental results shows that the proposed JFVO algorithm can converge in different communication states.}
\end{itemize}

% The rest of this paper is structured as follows: Section \ref{sec_non-IID_FL} explores and analyzes the FL performance with heterogeneous data. Section \ref{Sec:System} presents a clustered data sharing FL framework over wireless networks. Section \ref{Sec:Problem and Solution} formulates a problem aiming to minimize delays while mitigating the impact of non-IID data, and a joint optimization algorithm is designed to make a good tradeoff between the performance improvement and the cost. Section \ref{Sec_SimuRe} provides the numerical results, and finally conclusions are drawn in Section \ref{sec_Clusion}.

\section{Federated learning with non-IID data} \label{sec_non-IID_FL}
%In this section, we provide a clear definition of non-IID data in the context of federated learning, which is also referred to as "statistical heterogeneity". Then based on the classifications of non-IID data distributions, we apply the distribution distance to quantify the heterogeneity. Our preliminary experiments demonstrate the detrimental effects of non-IID data on the training of FedAvg, which is a typical federated learning algorithm, and a clustered data sharing method is proposed to address the non-IID challenge of FL without requiring an extra dataset.%

In this section, we provide a comprehensive definition of non-IID data in federated learning, which is also referred to as ``statistical heterogeneity''. We first introduce the classification of non-IID data distribution, and then apply the statistical distance to generally quantify the heterogeneity. Based on the preliminary experiments on the non-IID FL, a clustered data sharing method is proposed to address the non-IID challenge of FL without requiring extra datasets.
\subsection{Statistical Heterogeneity}
The local data generated at clients is statistically different, and may vary significantly, resulting in the typical non-IID cases. To illustrate this, consider a supervised learning task with input features $\mathbf{x}$ and labels $y \in \left\{ {1,...,Y} \right\}$. The statistical model of learning involves sampling the example-label pairs from the local data distribution of the client $k$, i.e., $\left( {{\mathbf{x}},y} \right) \sim {{P}_{k}}\left( {{\mathbf{x}},y} \right)$. If all local samples follow the same global distribution ${{P}_g}$ and are independent of each other, this is defined as IID in statistics, which has a mathematic form $\left( {{\mathbf{x}},y} \right) \sim {{P}_{g}}\left( {{\mathbf{x}},y} \right)$. Nevertheless, as users predominantly possess only localized data constrained by geo-regions and observation ability, the FL violates the IID assumption, resulting in distinctions between ${{P}_i}$ and ${{P}_j}$ for different clients $i$ and $j$.
We survey the following different cases for the non-IID assumption and rewrite ${{P}_k}({\mathbf{x}},y)$ as ${{P}_k}(y\left| {\mathbf{x}} \right.){{P}_k}({\mathbf{x}})$ or ${{P}_k}({\mathbf{x}}\left| y \right.){{P}_k}(y)$ allows us to characterize these different cases more precisely.

\textbf{\emph {Label distribution skew:}} The marginal distribution ${{{ P}}_k}(y)$ may vary across clients, even if ${{{ P}}_k}({\mathbf{x}}\left| y \right.)$ is the same, e.g., some labels only appear to a few users but not others.

\textbf{\emph{Feature distribution skew:}} The marginal distribution ${{{ P}}_k}({\mathbf{x}})$ may vary across clients, even if ${{{ P}}_k}(y\left| {\mathbf{x}} \right.)$ is the same, e.g., handwriting with the same label may vary in personal style.

Including the two common non-IID cases above, there are several other categories of distribution skews: \textbf{\emph{Concept shift}} refers to the same label ${{{ P}}_k}(y)$ with different features ${{{ P}}_k}({\mathbf{x}}\left| y \right.)$ or the same feature ${{{ P}}_k}({\mathbf{x}})$ with different labels ${{{ P}}_k}(y\left| {\mathbf{x}} \right.)$. \textbf{\emph{Quantity skew}} refers to the situation where the amount of data available for training differs significantly among participating clients.

Similar to the majority of non-IID FL works, our work also focuses on \emph{label distribution skew}, and can be extended to \emph{feature distribution skew} and other cases.

\subsection{Quantify Heterogeneity}
The evaluation of the distribution skewness is the first step for data owners and the FL server to explore the impact of non-IID data on training. The earth mover's distance (EMD) is generally applied as a metric to quantify non-IIDness between the local dataset and the global dataset. Specifically, given the local label distribution ${{{P}}_k}(y)$ and the global label distribution ${{{P}}_g}(y)$, the EMD of client $k$ can be calculated by
\begin{equation}
\label{Eq:define_EMD}
{D_{{\text{EMD}}}}\left( k \right) = \sum\limits_{i = 1}^Y {\left\| {{{P}_k}\left( {y = i} \right) - {{P}_{\text{g}}}\left( {y = i} \right)} \right\|} .
\end{equation}
There are also some other commonly used methods to measure the statistical distance, e.g., KL divergence \cite{KD_Data_FL} and JS divergence \cite{JS_Data_FL}. However, EMD is more widely applied in non-IID research \cite{FL_with_noniid,EMD1,EMD2}, because it can more accurately capture the impact of statistical heterogeneity on model training \cite{FL_with_noniid}.

Intuitively, the impact of non-IID data can be portrayed by the deviation between the model $w_g$ trained on the non-IID dataset and ${w_{{\rm{IID}}}}$ trained on the IID dataset. As model training is data-dependent, it can be inferred that the distance between data distributions can be used to measure the weight divergence $\left\| {{w_g} - {w_{{\rm{IID}}}}} \right\|$. Fortunately, some existing works have theoretically shown that ${D_{{\text{EMD}}}}$ is the root cause of the weight divergence \cite{FL_with_noniid,FL_on_noniid_Reinforce_learning}, i.e., $\left\| {{w_g} - {w_{{\text{IID}}}}} \right\| \propto {D_{{\text{EMD}}}}\left( k \right)$, and hence, $D_{{\text{EMD}}}\left( k \right)$ is an appropriate metric for non-IID FL. Moreover, we define the weighted sum of separate EMD values as the average EMD to reflect the overall data skewness:
\begin{equation}
\label{Eq:average_EMD1}
{{\bar D_{{\rm{EMD}}}} = \sum\limits_{k = 1}^K {\frac{{{n_k}}}{n}\sum\limits_{i = 1}^Y {\left\| {{{P}_k}\left( {y = i} \right) - {{P}_g}\left( {y = i} \right)} \right\|} }},
\end{equation}
where $n_k$ is the number of samples in client $k$, and $n = \sum\nolimits_{k = 1}^K {{n_k}} $ is the total number of samples across all $K$ clients. Note that the EMD can also be applied to quantify \emph{feature distribution skew} and other cases by replacing different distribution expressions.

\subsection{Preliminary Experiments and Analysis} \label{sec:pre_experiments}

To mitigate the divergence between the local and global distributions, one possible approach is to share data among the participating clients. Traditional data sharing schemes involve a centralized server sharing a portion of the global data, which is made available to all clients. However, this approach requires the creation of a publicly accessible proxy data source, which can be impractical or pose privacy concerns. Instead of centralized sharing, we randomly select clients to exchange local data within clusters. 
{The skewness of  the overall system is quantified using the average EMD ${\bar D_{\text{EMD}}}$. EMD is widely recognized as an effective metric for quantifying the statistical heterogeneity of data distributions in FL{\protect~\cite{FL_on_noniid_Reinforce_learning}}. To ensure fairness, the average EMD used in our work is weighted by the data volume of each user. Specifically, if users with significant deviations in data distribution have smaller dataset sizes, their influence on the average EMD is minimal. Conversely, users with larger dataset sizes and substantial distribution deviations exert a greater influence, resulting in a higher average EMD value.}
The preliminary experiments between this scheme and the standard FedAvg algorithm reveal that the convergence of the training process is impacted by three main factors.

\begin{figure*}[!t]
\centering
\subfloat[\hl{Impact of different non-IID degrees}]{\includegraphics[width=2.37in]{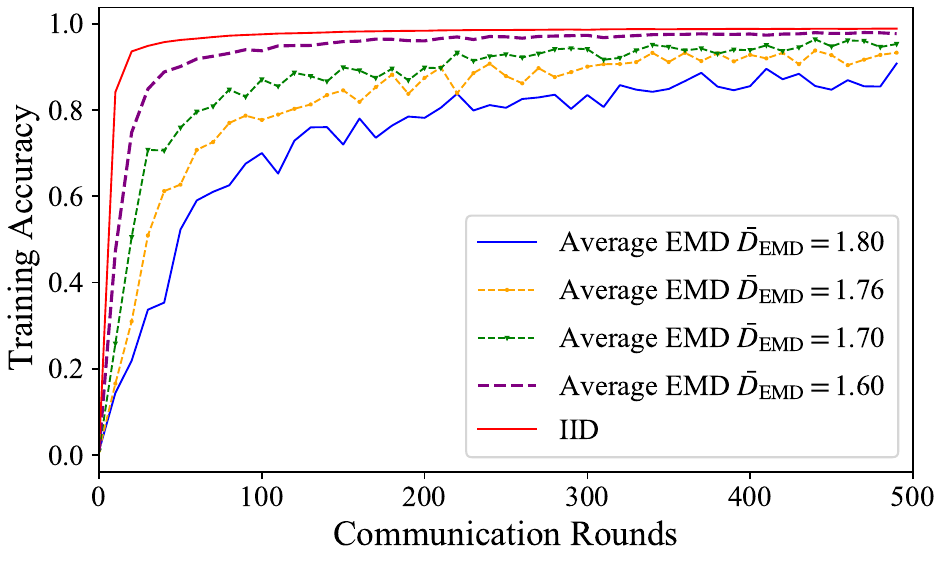}%
\label{fig1_1}}
\hfil
\subfloat[Impact of shared data volume]{\includegraphics[width=2.37in]{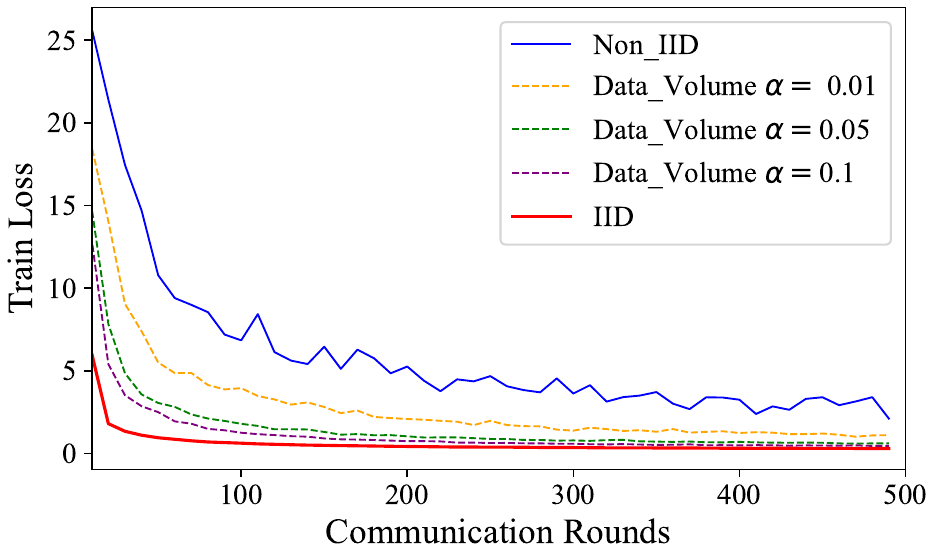}%
\label{fig1_2}}
\hfil
\subfloat[Impact of number of clusters]{\includegraphics[width=2.37in]{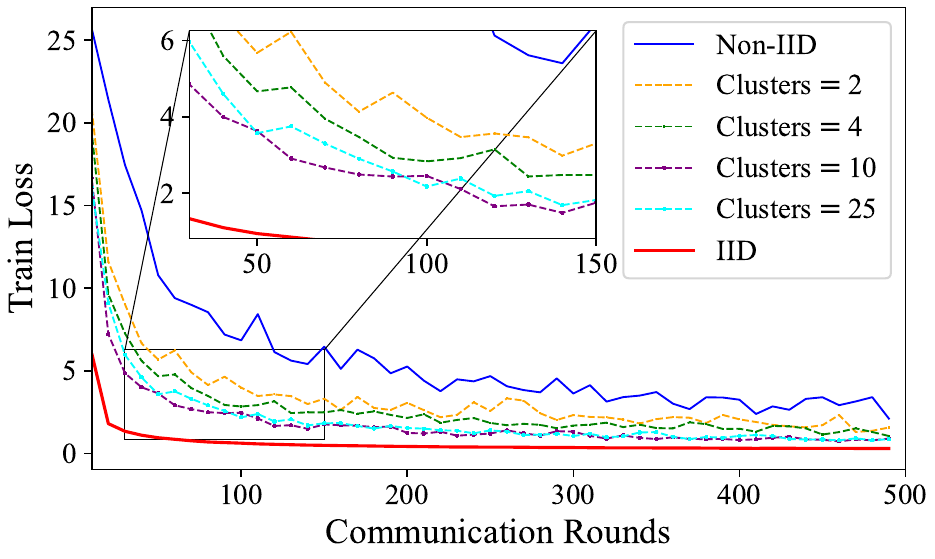}%
\label{fig1_3}}
\caption{Train loss on clustered data sharing for non-IID FL. (a) For IID setting, each user is randomly assigned a uniform distribution over all classes. For non-IID setting, the proportion of labels of among users are different, leading to varying ${\bar D_{{\text{EMD}}}}$\protect\footnotemark.  (b) The shared data volume is quantified as $\alpha$, representing the proportion of sharing data volume in its local data volume. (c) The users are randomly divided into clusters, where each cluster consists of cluster heads to transmit data and cluster members to receive data.}
\label{fig_sim}
\vspace{-5mm}
\end{figure*}
\footnotetext{{Please note that we assume all users have equal data volume, with only the proportion of labels changes, result in different average EMD.}} 

\emph{Degree of heterogeneity:}  {To further explore the relationship between data heterogeneity and FL training performance, we conduct experiments on training accuracy under varying $ \bar{D}_{\text{EMD}} $ settings in label distribution skew case. As shown in Fig. {\protect\ref{fig1_1}}, the results clearly indicate that reducing data heterogeneity significantly enhances the convergence rate of FL training, leading to higher model accuracy. These findings demonstrate that a higher $ \bar{D}_{\text{EMD}}$ slows the accuracy improvement per communication round, requiring more rounds to reach the same accuracy level as configurations with lower $ \bar{D}_{\text{EMD}} $.}

\emph{Shared data volume:} refers to how much data is transmitted to the clients. Sharing little data is ineffective for FL training, and much more data would increase the communication burden. To investigate the gains and costs of data sharing, we conduct the second experiment on clustered sharing with different shared data volumes.
As shown in Fig. \ref{fig1_2}, as the volume of shared data increases, the convergence rate of FL also increases. Furthermore, it is worth noting that the convergence improvement achieved by raising the shared data volume from $\alpha=0.05$ to $\alpha=0.1$ is not as significant as from $\alpha=0$ (without data sharing) to $\alpha=0.05$.

\emph{Number of clusters:} plays a critical role in determining the effectiveness of the data sharing scheme. Increasing the number of clusters can reduce transmission costs but may not be beneficial for FL training. As shown in Fig. \ref{fig1_3}, it can be observed that excessive clustering can have a negative impact on FL performance instead of improving it when comparing the cases of clusters $=10$ with clusters $=25$. Too many clusters may result in data sharing being inefficient, and random data sharing itself can not always guarantee performance gains and can incur communication costs.

Based on the experiments and analyses presented above, several notable conclusions can be drawn:
\begin{enumerate}
\item{The lower skewness datasets converge faster during training and achieve higher accuracy, which can be characterized by the lower average-EMD between the data distributions.}
\item{The benefits of data sharing increase as the shared data volume rises, but the rate of improvement is diminishing. In particular, when the transmitter shares an excess of data, the cost of data sharing becomes the critical bottleneck for FL training instead of heterogeneity. Therefore, it is important to make the trade-off between data sharing and FL training.}
\item{The effectiveness of clustered data sharing depends on the clustering strategy. Combined with item 1 above, it is clear that an efficient clustering method for accelerated non-IID FL does not depend on the number of clusters but on the contribution to the average EMD ${\bar D_{{\rm{EMD}}}}$.}
\end{enumerate}

In conclusion, it is critical to design a reasonable clustering method to accelerate federated learning with acceptable sharing delay, i.e., minimizing the total delay of data sharing and FL training.

\section{System Model} \label{Sec:System}
Consider a wireless edge computing system consisting of a BS and a set of $K$ users (i.e., smart mobile devices) denoted as ${\cal K} = \{ 1,2,...,K\} $. Each user collects the local samples ${{{\cal N}}_k} = \{ {\bf{x}}_i^k,y_i^k\} _{i = 1}^{{n_k}},\;k \in {{\cal K}}$, where ${\bf{x}}_i^k$ is the input vector of sample $i$, $y_i^k$ is the corresponding labels, and ${n_k}$ is the number of the sample-label pairs.  The collaborative interaction between users and the BS is facilitated within a client-server architecture to accomplish intelligent services, e.g., computer vision tasks. In this edge environment, besides the conventional downlink/uplink transmission between clients and the server, sidelink-assisted multicasting are enabled for users in close proximity.
\begin{figure*}[t]
\centering
\includegraphics[width=13.8cm,height=5.8cm]{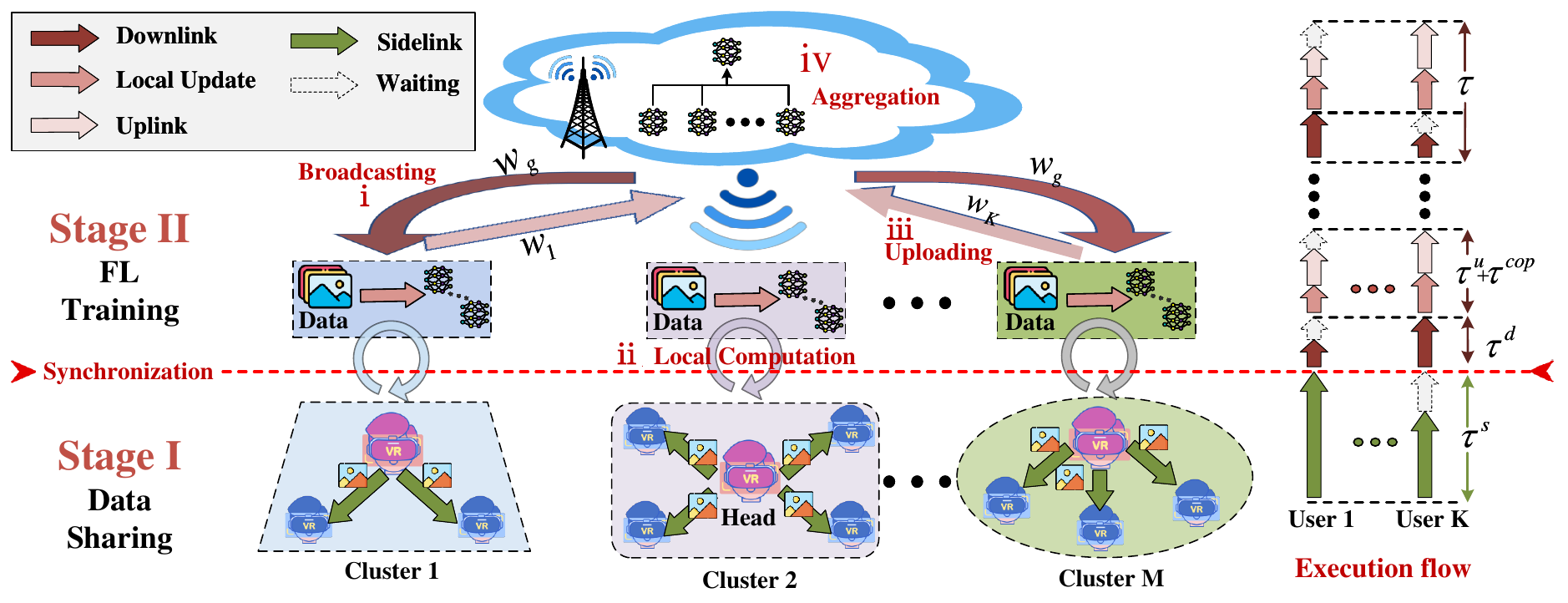}
\vspace{-3mm}
\caption{Clustered data sharing framework for FEL.}
\vspace{-5mm}
\label{fig2}
\end{figure*}

\subsection{Clustered Data Sharing FEL Framework} \label{sec:Framework}
To relieve the data imbalance feature for FL, we propose a communication-aware clustered data sharing scheme which makes data distributions more homogeneous among devices by exchanging a small data subset within a communication-efficient and privacy-preserving cluster. As shown in Fig. \ref{fig2}, this framework consists of the following two stages.

\textbf{The data preparing stage:} involves a data augmentation procedure before the FEL training. In this stage, all users are clustered with respect to data distribution, communication states, and privacy requirements, where each cluster is defined by a cluster $m$ and its corresponding cluster members ${{\cal C}_m}$. Then the cluster heads, denoted as the set ${\cal M}$, share the subsets of their data with cluster members through reliable multicast communication. Following clustered data exchange, each node can compute its local parameters with the mixed samples in the next stage. It is noted that each node cannot participate in more than one cluster to prevent conflicts, i.e., ${{\cal C}_m} \cap {{\cal C}_l} = \emptyset ,\forall m,l \in {\mathcal M},m \ne l$. With a copy of shared data, the local data volume on device $k$ is then determined by the following expression:
\begin{equation}\label{Eq:local_data}
	{{\tilde n_k} = \left\{ \begin{array}{l}
{n_k} + n_m^s, \; \; k \in {{\cal C}_m}, \; \forall m \in \mathcal{M},\\
{n_k},\quad\quad\quad\,{\rm{      }}k \notin {{\cal C}_m}, \; \forall m \in \mathcal{M},
\end{array} \right.}
\end{equation}
where $n_m^s$ is the data volume of sharing from cluster head $m$. Without loss of generality, we specify ${P_m}\left( y \right) = {P_k}\left( y \right)$ if $k \notin {{\cal C}_m}$. The new data distribution on device $k$ can be derived as follows:
\begin{equation}\label{Eq:local_distribution}
	{{\tilde P_k}\left( y \right) = \frac{{{n_k}{P_k}\left( y \right) + n_m^s{P_m}\left( y \right)}}{{{n_k} + n_m^s}}}.
\end{equation}
Intuitively, clustered data exchange alters the original data distribution on each device, diluting the disparity in data features. Therefore, the value of the average EMD after data sharing is given by
\begin{equation}\label{Eq:averege_EMD2}
	{{\tilde D_{{\rm{EMD}}}} = \sum\limits_{k = 1}^K {\frac{{{{\tilde n}_k}}}{n}\sum\limits_{i = 1}^Y {\left\| {{{\tilde P}_k}\left( {y = i} \right) - {P_g}\left( {y = i} \right)} \right\|} } }.
\end{equation}

\textbf{The FEL training stage:} is to train a global model via collaborations between the edge server and multiple clients, where the global model is obtained by minimizing the following loss function:
\begin{equation}\label{Eq:loss_func}
    {\mathop {\min }\limits_{{w_g}} \frac{1}{K}\sum\limits_{k = 1}^K {\frac{1}{{{n_k}}}\sum\limits_{i = 1}^{{n_k}} {{F_k}\left( {{\bf{x}}_i^k,y_i^k,{w_g}} \right)} } }.
\end{equation}
Here, ${F_k}\left(  \cdot  \right)$ represents the loss of prediction on the local sample $\left\{{\bf{x}}_i^k,y_i^k \right\}$. Specifically, the model undergoes the following steps for updating during the $t$-th round:

\emph{i) Global Model Broadcast:} The server broadcasts the global model ${w_g^{t - 1}}$ to participants, updating their local models to ${w_k^t}$.

\emph{ii) Local Model Update:} Participants independently train their local models $w_k^t$ using their respective datasets $\left\{ {{\bf{x}}_i^k,y_i^k} \right\}_{{\rm{i = 1}}}^{{n_k}}$ via stochastic gradient descent (SGD) \cite{SGD}:
$ {w_k^t} \leftarrow {w_k^t} - \eta \nabla {F_k}\left( {{x_i},{y_i};{w_k^t}} \right) $
where $\eta $ is the learning rate in SGD, and $\nabla {F_k}\left(  \cdot  \right)$ is the gradient of the loss function ${F_k}\left(  \cdot  \right)$ with respect to $w_k^t$.

\emph{iii) Parallel Upload:} After local updates, participants transmit $w_k^t$ to the BS via wireless cellular links.

\emph{iv) Synchronous Aggregation:} Once all participants have completed local model transmission, the global model $w_g^t$ is updated by weighted-averaging the uploaded models:
\begin{equation}\label{Eq:aggregation}
    {w_g^t = \sum\limits_{k = 1}^K {\frac{{{n_k}}}{n}w_k^t} }.
\end{equation}

After model aggregation, $w_g^t$ is fed back to the devices, and the FEL above training procedure repeats for multiple rounds until convergence.

\subsection{Clustered Data Sharing Model}
As is well-known, data sharing is inherently associated with privacy risks and additional communication costs. To effectively mitigate these effects, two key metrics are defined for the further development and refinement of the cluster algorithm.

\emph{Social closeness:} is one of the factors of social awareness, representing users with different levels of trust in each other \cite{socail_closeness}. Regarding data privacy, users' reluctance to expose data is weakened in close social relationships. For clustered data sharing, we define the social closeness ${e_{m,c}}$ between the cluster head $m \in M$ and cluster member $c \in {{\cal C}_m}$. To ensure trustworthy data exchange, cluster members are required to maintain a socially closed affiliation with the cluster head, i.e., ${e_{m,c}} \ge {e_{{\rm{th}}}},\forall m,c$. For clarity, we build a graph ${{\cal G}} = \left( {{{\cal K}},{{\cal E}}} \right)$ to illustrate the social closeness among users, where ${{\cal E}} = \{ {e_{k,j}} \in \left[ {0,1} \right],\forall k,j \in {{\cal K}}\} $ represents the set of graph edges \cite{D2D_Social_Aware}.

\emph{Multicasting delay:} The cluster head shares part of the datasets with the group members via sidelink-assisted multicasting. Sidelink is a technology in 5G/6G allowing for Device-to-Device (D2D) communication without using BS \cite{Muticasting_Survey}.
%Following 3GPP standard \cite{3GPP_6GChannel}, the path loss model for urban microcell (UMi) can be defined as:
%\begin{equation}\label{Eq:path_loss}
%	{\text{PL}\left( {{d_{m,c}}} \right)[dB] = \beta_{PL} + 10\zeta \log _{10}{{d_{m,c}}} + 20\log _{10}{f_c^s}},
%\end{equation}
%where $d_{m,c}$ denotes the three-dimensional (3D) distance between cluster head $m$ and member $c$, and $f_c^s$ represents the carrier frequency of data sharing in GHz. The coefficients $\beta_{PL}$ and $\zeta$ are contingent upon the environment, e.g., the line-of-sight (LoS) and non-line-of-sight (nLoS) states, as well as the blocked and non-blocked channel conditions.
Then the general formulation of Signal to Interference and Noise Ratio (SINR) $\gamma$ is given by
\begin{equation}\label{Eq:SNR_Multicasting}
	{\gamma\left( {{d_{m,c}}} \right) = \sum\limits_{i = 0}^3 {\frac{{{C_i}A_i^{ - 1}d_{m,c}^{ - {\zeta _i}}{\kappa _i}\left( {{d_{m,c}}} \right){S_{F,i}}{U_{F,i}}}}{{{N_0}{B_m^s} + I}}}},
\end{equation}
where ${C_i} = {P_m}{G_m}{G_c}$, and $P_m$ is the transmit power, $G_m$ and $G_c$ are the antenna array gains at cluster head $m$ and member $c$. $d_{m,c}$ denotes the three-dimensional (3D) distance between cluster head $m$ and member $c$. ${\kappa _i},i \in \left\{ {0,1,2,3} \right\}$ is the state probabilities of four states: non-line-of-sight (nLoS) with blocked, line-of-sight (LoS) with blocked, nLOS with non-blocked, LOS with non-blocked. $A_i$ and $\zeta_i$ are propagation coefficients of the path loss, which corresponding to the introduced states. ${S_{F,i}}$ and ${U_{F,i}}$ are state-dependent shadow and small-scale fading. $N_0$ is the noise power spectral density, $B_m^s$ is the bandwidth of multicasting in cluster $m$ and $I$ is the interference. In the urban microcell (UMi) scenario, we set these coefficient values according to the 3GPP standards \cite{3GPP_6GChannel}.
%where $P_m$ and $D_m$ are the transmit power and the antenna gain of cluster head $m$. $N0$ is the power spectral density of the Gaussian noise. $B_m^s$ is the multicasting bandwidth, $M_I$ and $M_S$ are interference and shadow fading margins.
With multicasting, the transmission rate of data sharing within one cluster is determined by the worst sidelink, i.e.,
%The transmission rate of data sharing from cluster head $m$ to each corresponding member $c$ is given by
%\begin{equation}\label{Eq:D2D_rate}
%	{{v_{m,c}} = B_{m,c}^s{\log _2}\left( {1 + \frac{{P_m^sh_{m,c}^2}}{{{I_m^s} + B_{m,c}^s{N_0}}}} \right)},
%\end{equation}
%where $B_{m}^s$ is the bandwidth used by cluster head $m$ for multicasting the partial dataset; ${h_{m,c}}$ is the channel gain between cluster head $m$ and member $c$. $P_m^s$ is the transmit power of cluster head, and ${N_0}$ is the power spectral density of the Gaussian noise. ${I_m^s}$ is the inference caused by other clusters located in different service areas. With multicast, the delay for data sharing within one cluster is determined by the worst sidelink, i.e.,
\begin{equation}\label{Eq:D2D_cluster_delay}
	{v_m^s = \mathop {\min }\limits_{c \in {\mathcal{C}_m}} \left\{ {B_m^s{{\log }_2}\left( {1 + \gamma \left( {{d_{m,c}}} \right)} \right)} \right\}},
\end{equation}
Meanwhile, the total data sharing time cost ${\tau ^s}$ depends on the maximal delay of clusters, and is thereby given by
\begin{equation}\label{Eq:D2D_delay}
	{{\tau ^s} = \mathop {\max }\limits_{m \in \mathcal{M}} \left\{ {\frac{{an_m^s}}{{v_m^s}}} \right\}},
\end{equation}
where $a$ denotes bits of each data sample. The energy consumption of one-shot data sharing can be ignored compared to the multi-round FEL training process.

\subsection{FEL Transmission and Computation Model}
As introduced in Section \ref{sec:Framework}, the FEL training stage comprises four steps in each round, with each step incurring either communication or computational cost.

\emph{i) Broadcasting:}
The downlink transmission rate of the BS when broadcasting the global model to user $k$ can be expressed as follows:
\begin{equation}\label{Eq:Download_rate}
	{v_k^d = {B^d}{\log _2}\left( {1 + \frac{{{P_{{\text{BS}}}} {\left| h_k \right|}^2}}{{{I^d} + {B^d}{N_0}}}} \right)},
\end{equation}
where ${B^d}$ is the bandwidth used by the BS to broadcast to global model; ${P_{{\text{BS}}}}$ is the transmit power of the BS, and ${h_k}$ is the channel gain between the BS and user $k$; $I^d$ represents the interference caused by other BSs which are not participating in training. As the channel state fluctuates during the downloading process, maintaining a constant transmission rate is infeasible. Thus, we employ the expected value of the transmission time cost to represent the true value. The corresponding downloading delay can be expressed as
\begin{equation}\label{Eq:Download_delay}
	{\bar \tau _k^d = {\mathbb{E}_{{h_k}}}\left( {\frac{{M\left( {{w_g}} \right)}}{{v_k^d}}} \right)},
\end{equation}
where function $\mathbb{E}_{{h_k}} \left( \cdot \right)$ is the expectation with respect to $h_k$ and function $M\left( \cdot \right)$ is the model size of model $w_g$ in bits.

\emph{ii) Local computation:}
Assuming that $f_k$ denotes the computation capacity of user $k$ (measured in CPU cycles per second),
The computation latency at user $k$ needed for data processing can be expressed by
\begin{equation}\label{Eq:Computation_delay}
	{\tau _k^{cop} = \frac{{{L_k}E{{\tilde n}_k}}}{{{f_k}}}}.
\end{equation}
Here, $L_k$ denotes the number of CPU cycles needed for training per sample at user $k$. $E$ is the number of local epochs used in SGD updateing, indicating the number of times the local device iterates over its own subset of training data ${\tilde n}_k$. The energy consumption of each user $k$ for local model training can be expressed by
\begin{equation}\label{Eq:Computation_energy}
	{\gamma _k^{cop} = {\varsigma}{L_k}E{{\tilde n}_k}f_k^2},
\end{equation}
where $\varsigma$ is the energy consumption coefficient depending on the hardware of the devices.

\emph{iii) Uploading:}
After local computation, all users upload their local model parameters to the BS via orthogonal frequency division multiple access (OFDMA). The achievable uploading rate of user $k$ can be expressed as
\begin{equation} \label{Eq:upload_rate}
{v_k^u = {B^u}\sum\limits_{i = 1}^R {{r_{i,k}}} {\log _2}\left( {1 + \frac{{{P_k}{\left| h_k \right|}^2}}{{I_k^u + B_k^u{N_0}}}} \right)},
\end{equation}
where $B_k^u$ is the channel bandwidth assigned to user $k$,  ${\mathbf{R}} = \left\{ {{r_{i,k}}\left| {{r_{i,k}} \in \left\{ {0,1} \right\},i \in R,k \in \mathcal{K}} \right.} \right\}$ denotes the subcarrier allocation matrix, and $r_{i,k} \hspace{-1.2mm} = \hspace{-1.2mm}1$ means the subcarrier $i$ is allocated to user $k$, and $r_{i,k} \hspace{-1mm} = \hspace{-1mm}0$ means not. $P_k$ is the transmission power of user $k$, $h_k$ is the channel gain between user $k$ and the BS, and $I_k^u$ is the interference caused by the devices using the same subcarriers for other services. The corresponding uploading delay can be expressed as
\begin{equation} \label{Eq:upload_delay}
{\bar \tau _k^u = {\mathbb{E}_{{h_k}}}\left( {\frac{{M\left( {{w_k}} \right)}}{{v_k^u}}} \right)},
\end{equation}
and the uploading energy consumption can be calculated as
\begin{equation} \label{Eq:upload_energy}
{\gamma _k^u = {P_k}\tau _k^u}.
\end{equation}

\emph{iv) Aggregation:}
Since the BS has considerable computing capability and a continuous power supply, the delay and energy consumption during aggregation can be ignored.

According to the system model, the FEL latency is composed of three parts: downloading, computation, and uploading delay, and all clients have to complete the synchronization of the broadcast before proceeding with any local update. Hence, the total delay of one FEL round with the users and the BS jointly participating can be expressed as
\begin{equation} \label{Eq:total_dely}
{\tau  = \mathop {\max }\limits_{k \in \mathcal{K}} {\text{ }}\left\{ {\bar \tau _k^d} \right\} + \mathop {\max }\limits_{k \in \mathcal{K}} \left\{ {\tau _k^{cop} + \bar \tau _k^u} \right\}}.
\end{equation}
Due to the fact that communication rounds are independent, it suffices to consider the energy consumption for an arbitrary round without loss of generality \cite{Energy_Efficient_FL}. Therefore, the total energy consumption for a single FEL round, considering both computation and uploading energy, can be expressed as
\begin{equation}\label{Eq:total_energy}
{{\gamma _k} = \gamma _k^{cop} + \gamma _k^u}.
\end{equation}

\section{Problem Formulation and Solution} \label{Sec:Problem and Solution}
The critical role of an appropriate clustering method in wireless federated learning training is highlighted in Section \ref{sec:pre_experiments}, emphasizing its significance in achieving optimal performance. We formulate an optimization problem by jointly considering the communication, computation, and FL performance in this clustered data sharing framework. Subsequently, the original problem can be solved by decomposing it into two subproblems and insights into clustering are provided by the detailed analysis of data sharing on data heterogeneity.
\subsection{Problem Formulation and Decomposition}
To maximize the effectiveness of data sharing in accelerating FEL training, we formulate an optimization framework to minimize the total delay for training the whole FEL algorithm. This minimization problem joints the design of the clustering strategy and resource optimization (i.e., shared data volume and computed frequency) for each cluster while guaranteeing  privacy as follows:
\addtocounter{equation}{0}
\begin{equation}\label{eq:Pri_problem}
\begin{split}
\!\!\!\!\!\!\!\!\!\!{\mathcal{P}0:} \mathop {\min }\limits_{\mathcal{M},\mathcal{C},\boldsymbol{F},{{\boldsymbol{N}}^s}} \; {\tau ^s} + \sum\limits_{t = 1}^T \tau ,\quad\quad\quad\quad\quad\quad\quad
\end{split}
\end{equation}
\vspace{-0.5cm}
\begin{align}
&\!\!\!\!\!\!\!\!\!\! \rm{s.\;t.} \scalebox{1}{$\;\;\;\; \theta  \ge {\theta _{{\text{th}}}},$} \tag{\theequation{a}} \label{C_1_a}\\
&\scalebox{1}{$\;\;\;\; {\mathcal{C}_m} \cap {\mathcal{C}_l} = \emptyset ,\;\;\forall m,\;l \in \mathcal{M},\;m \ne l, $} \tag{\theequation{b}} \label{C_1_b}\\
&\scalebox{1}{$\;\;\;\;  {e_{m,c}} \ge {e_{{\text{th}}}},\;\;\forall m \in \mathcal{M},\;c \in {\mathcal{C}_m}, $} \tag{\theequation{c}} \label{C_1_c}\\
&\scalebox{1}{$\;\;\;\;  {\text{0}} \le n_m^s \le {n_m}, \;\; \forall m \in \mathcal{M},$} \tag{\theequation{d}} \label{C_1_d}\\
&\scalebox{1}{$\;\;\;\; 0 \le {f_k} \le {f_{\max }}, \;\; \forall k \in \mathcal{K}, $} \tag{\theequation{e}} \label{C_1_e}\\
&\scalebox{1}{$\;\;\;\;  {\gamma _k} \le {\gamma _{{\text{th}}}},\;\;\forall k \in \mathcal{K},$} \tag{\theequation{f}} \label{C_1_f}
\end{align}
where ${{\cal C}}{\rm{ = }}\left\{ {\left. {{{{\cal C}}_1} \ldots ,{{{\cal C}}_m}, \ldots } \right\}} \right.$ and ${{{\cal C}}_m}$ is the set of cluster members connected with the cluster head $m$. $\boldsymbol{F} = \left[ {{f_1}, \ldots ,{f_K}} \right]$ is a computed frequency vector with $K$ being the total number of devices. ${\boldsymbol{N}}^s = \left[ {n_1^s, \ldots ,n_m^s, \ldots } \right]$ is a vector of sharing data volume from the cluster heads. $T$ is the number of global round needed to attain the target accuracy $\theta _{{\text{th}}}$. Constraint (\ref{C_1_a}) is the preset accuracy requirement for FL training. (\ref{C_1_b}) is the constraint for exclusive cluster forming. Constraint (\ref{C_1_c}) restricts the privacy requirement for data sharing between cluster heads and cluster members. The shared data volume should satisfy constraint (\ref{C_1_d}). $f_{\max}$ is the maximum local computation capacity, and constraint (\ref{C_1_e}) presents the frequency limits of all devices. Constraint (\ref{C_1_f}) is the energy consumption requirement of FEL training at each round. Accordingly, the multi-constraints involved objective achieve the joint optimization of communication, computation and learning performance by exploiting tradeoff of the data and computility resource at the edge.

As shown in (\ref{eq:Pri_problem}), there are several difficulties in directly solving the primal non-convex optimization problem. \emph{Firstly}, the relationship between clustering strategy, shared data volume, and communication round $T$ is implicit due to the no-closed expression. \emph{Secondly}, the joint decision of optimization variables is coupled, and the clustering and cluster head selection construct an NP-hard problem that existing practices all rely on heuristic algorithms. \emph{Thirdly}, the cluster formation incorporates restrictions on privacy preservation and transmission efficiency as well, making the problem further intricate.

In order to solve problem $\mathcal{P}0$, we adopt a decomposition approach by recognizing that the FEL training process can only commence once the clustering process has been completed. As such, we break down the primal problem into the following two subproblems, based on the sequential order of the clustering stage and the FEL training stage.

\emph{Subproblem 1: Efficient Cluster Forming Strategy with Precision and Privacy Constraints.} The optimal clustering strategy is designed to encompass the entire clustering and FEL training stages, with the objective of minimizing the overall delay, i.e.,
\addtocounter{equation}{0}
\begin{equation}\label{eq:sub_problem_1.0}
\begin{split}
\!\!\!\!\!\!\!\!\!\!{\mathcal{P}1:} \mathop {\min }\limits_{\mathcal{M},\mathcal{C}} \; {\tau ^s} + T \cdot \tau ,\quad\quad\quad\quad\quad\quad\quad
\end{split}
\end{equation}
\vspace{-0.5cm}
\begin{align}
&\!\!\!\!\!\!\!\!\!\! \rm{s.\;t.} \scalebox{1}{$\;\;\;\; \theta  \ge {\theta _{{\text{th}}}},$} \tag{\theequation{a}} \label{sub1.0_C__a}\\
&\scalebox{1}{$\;\;\;\; {\mathcal{C}_m} \cap {\mathcal{C}_l} = \emptyset ,\;\;\forall m,\;l \in \mathcal{M},\;m \ne l, $} \tag{\theequation{b}} \label{sub1.0_C_b}\\
&\scalebox{1}{$\;\;\;\;  {e_{m,c}} \ge {e_{{\text{th}}}},\;\;\forall m \in \mathcal{M},\;c \in {\mathcal{C}_m}. $} \tag{\theequation{c}} \label{sub1.0_C_c}
\end{align}

Although decoupling diminishes the complexity of directly solving the primal problem, the subproblem $\mathcal{P}1$ remains no-trivial due to the lack of a closed-form between objective function and variables. The training delay (right term), influenced by the multi-round nature of the process, constitutes the dominant factor in the total delay. Consequently, we simplify the optimization objective as the right term, and the data sharing time $\tau ^s$ as a additional constraint.  As depicted in Figure \ref{fig1_1}, a smaller average EMD requires fewer communication rounds $T$, which can achieve target accuracy with less training delay. Thus, the right term of objective in $\mathcal{P}1$ is equivalent to minimize the average EMD of all local distributions, i.e., (\ref{Eq:averege_EMD2}). Concurrently, given the known shared data volume, the data sharing time, calculated by (\ref{Eq:D2D_delay}), depends on the transmission rate of data sharing. Thus a proper sharing preparing duration is ensured by an additional rate threshold $v_{\rm{th}}$.

\emph{Subproblem 2: Computed Frequency and Shared Data Volume Allocation Subproblem.} During the training stage, the computed frequency and shared data volume are jointly optimized to minimize the overall delay, i.e.,
\addtocounter{equation}{0}
\begin{equation}\label{eq:Sub_problem2}
\begin{split}
\!\!\!\!\!\!\!\!\!\!{\mathcal{P}2:} \mathop {\min }\limits_{\boldsymbol{F},{{\boldsymbol{N}}^s}} \; {\tau ^s} + T \cdot \tau ,\quad\quad\quad\quad\quad\quad\quad
\end{split}
\end{equation}
\vspace{-0.5cm}
\begin{align}
&\!\!\!\!\!\!\!\!\!\! \rm{s.\;t.} \scalebox{1}{$\;\;\;\;  {\text{0}} \le n_m^s \le {n_m}, \;\; \forall m \in \mathcal{M},$} \tag{\theequation{a}} \label{SUB2_C_a}\\
&\scalebox{1}{$\;\;\;\; 0 \le {f_k} \le {f_{\max }}, \;\; \forall k \in \mathcal{K}, $} \tag{\theequation{b}} \label{Sub2_C_b}\\
&\scalebox{1}{$\;\;\;\;  {\gamma _k} \le {\gamma _{{\text{th}}}},\;\;\forall k \in \mathcal{K}.$} \tag{\theequation{c}} \label{SUB2_C_c}
\end{align}
This subproblem is challenging since it involves a non-convex optimization problem with no closed-form objective (the function of $T\left( {{\boldsymbol{N}}^s} \right)$ is unknown). We first estimate the function $T\left( \cdot \right)$ based on the experimental data, and then this subproblem can be transformed into a stochastic optimization problem with uncertain parameters.

%Based on the aforementioned analysis, the complexity of the problem mainly focus on the clustering decision and the closed expression of communication rounds $T$. As the findings in \ref{sec_non-IID_FL}, the implementation of clustered sharing can effectively reduce communication consumption in FL by minimizing data heterogeneity among clients. Thus, there is another optimization direction toward designing clustering strategy, i.e., minimizing the average EMD of all local distribution with the fixed shared data volume and computed frequency. We can estimate the communication round $T$ function by fitting data from a set of FL experiments. For problem decomposition, we divide the primal problem into two subproblem, i.e., \emph{the cluster heads, members selection} and \emph{computed frequency, shared data volume optimization}.

\subsection{Distribution-based Adaptive Clustering Algorithm} \label{Sec_DACA}
The cluster heads and members selection should be optimized to minimize the average EMD while ensuring user privacy and high communication rates as follows:
\addtocounter{equation}{0}
\begin{equation}\label{eq:Sub_problem1}
\begin{split}
{\mathcal{P}1':} \;\;\;\;\mathop { \min }\limits_{{\cal M},{{\cal C}}} \;\;\; {\tilde D_{{\rm{EMD}}}},\quad\quad\quad\quad\quad\quad\quad\quad\quad\quad
\end{split}
\end{equation}
\vspace{-0.8cm}
\begin{align}
&\!\!\!\!\!\!\!\!\!\! \rm{s.\;t.} \scalebox{1}{$\;\;\;\; \theta  \ge {\theta _{{\text{th}}}},$} \tag{\theequation{a}} \label{C_2_a}\\
&\scalebox{1}{$\;\;\;\; {\mathcal{C}_m} \cap {\mathcal{C}_l} = \emptyset ,\;\;\forall m,\;l \in \mathcal{M},\;m \ne l, $} \tag{\theequation{b}} \label{C_2_b}\\
&\scalebox{1}{$\;\;\;\; {e_{m,c}} \ge {e_{\text{th}}},\;\;\;\forall m \in {{\cal M}},\;c \in {{\cal C}}_m ,$} \tag{\theequation{c}} \label{C_2_c}\\
&\scalebox{1}{$\;\;\;\; {v_{m,c}} \ge {v_{\text{th}}},\;\;\;\forall m \in {{\cal M}},\;c \in {{\cal C}}_m ,$} \tag{\theequation{d}} \label{C_2_d}
\end{align}
where the lower bound of transmission rate $v_{th}$ maintain a proper sharing utility. The data sharing strategy for tuning the non-IID degree is powerful, yet intractable because the variables of $\mathcal{M}$ and $\mathcal{C}$ are coupled and can affect the objective value of $\tilde{D}_{\text{EMD}}$. \emph{To identify the complex clustering effect on} $\tilde{D}_{\text{EMD}}$, we set aside the external constraints in (\ref{C_2_c}) and (\ref{C_2_d}). Next, we devise the following three conditions to analyze the reward of data sharing on $\tilde{D}_{\text{EMD}}$ from the perspectives of the individual, intra-cluster, and inter-cluster.

\textbf{Condition 1: (individual perspective)}
After data sharing, the EMD of an arbitrary user $k$ should be as small as possible, i.e.,
%\vspace{-4mm}
\begin{equation}\label{Eq:condition_1}
	{\mathop {\min }\limits_{\forall k \in {{\cal K}}} \sum\limits_{i = 1}^Y {\left\| {{{\tilde P}_k}\left( {y = i} \right) - {P_g}\left( {y = i} \right)} \right\|} }.
\end{equation}
Especially if {\small $\sum\nolimits_{i = 1}^Y {\left\| {{{\tilde P}_k}\left( {y = i} \right) - {P_g}\left( {y = i} \right)} \right\|} \hspace{-1mm} =\hspace{-1mm} 0$}, the data distribution turns into the ideal IID. Without considering any constraints, condition 1 is equivalent to the original optimization problem. But it is still hard to meet this condition, and thus we convert the target into the following two extended conditions:

%\addtolength{\topmargin}{0.04in}
\textbf{Condition 2: (intra-cluster perspective)}
Within a cluster, the best data sharing is secured if the EMD of the cluster head and cluster members differs as much as possible, i.e.,
%\vspace{-3mm}
\begin{equation}\label{Eq:condition_2}
	{\mathop {\max }\limits_{\forall c \in {{\cal C}_m}} {D_{{\rm{EMD}}}}\left( c \right) - {D_{{\rm{EMD}}}}\left( m \right)}.
\end{equation}
Regarding the EMD of cluster members is already given, to release the divergence after data sharing, the EMD of cluster heads should be as low as possible. In brief, nodes with high data quality are preferred to be selected as cluster heads. While condition 2 guarantees optimal intra-cluster data sharing, the appropriate number of clusters and inter-cluster relationships are still undetermined. Hence, an additional constraint is necessary to define inter-cluster relationships.

\textbf{Condition 3: (inter-cluster perspective)}
Given the results of a particular clustering ${\cal M}$ and ${\cal C}$, there exists a better clustering method if the distribution distance of any member $c \in {{\cal C}_m}$ and other cluster head $m'$ is larger than that with the current head $m$, i.e.,
%\vspace{-3mm}
\begin{equation}\label{Eq:condition_3}
%{\mathop {\max }\limits_{m \in {\cal M}} {D'_{{\rm{EMD}}}}\left( {k,m} \right)}.
{{D_{{\rm{EMD}}}}\left( {c,m} \right) \ge {D_{{\rm{EMD}}}}\left( {c,m'} \right),\forall c \in {{\cal C}_m},m \ne m'}.
\end{equation}
Here,{\small{ ${D_{{\rm{EMD}}}}\left( {c,m} \right)\hspace{-1mm}={D_{{\rm{EMD}}}}\left( {c} \right) - {D_{{\rm{EMD}}}}\left( {m} \right)$}} is the difference in EMD between cluster member $c$ and the cluster head $m$. For nodes that have a critical role and can potentially belong to multiple clusters, this condition can be utilized as a separation criterion. An intuitive interpretation is that these nodes will be assigned to the cluster whose data distribution of the cluster head differs significantly from their own.

%\begin{remark}
%  \cite{14} groups the devices into multiple clusters that perform FL cyclically in each round and the convergence rate depends on the data heterogeneity of %intra-cluster data.
%\end{remark}

%\begin{remark}
%  \cite{12,17} trains the multi-models in clustering users by aggregating the local models from the same cluster. In this case, the data heterogeneity is from the inter-cluster data.
%\end{remark}

Aided by the condition 2 and the condition 3, the solutions to the relaxed problem $\mathcal{P}1$ can be obtained by the following theorem.

\begin{theorem} \label{thm}
\emph{Putting the constraints (\ref{C_2_c}) and (\ref{C_2_d}) aside, if ${{\cal M}^*}$ and ${{\cal C}^*}$ are the optimal solutions to  the problem $\mathcal{P}1'$, then condition 2 and condition 3 both hold with ${{\cal M}^*}$ and ${{\cal C}^*}$.}
%\emph{If condition 2 and condition 3 both hold with ${{\cal M}^*}$ and ${{\cal C}^*}$, then ${{\cal M}^*}$ and ${{\cal C}^*}$ are the optimal solutions of problem %(\ref{Eq:11}).}

The proof is omitted in this paper due to the limited space. The full proofs can refer to Appendix or online version \cite{Faster_FL_arxiv}.
\end{theorem}

By integrating constraints (\ref{C_2_c}) and (\ref{C_2_d}), which involve transmission rates and privacy thresholds among the associates, problem $\mathcal{P}1'$ is impacted, making Theorem \ref{thm} not immediately applicable. These constraints are associated with connectability with credible and communication-efficient nodes. Following the preceding analysis, we reformulate a constrained graph as {\small ${\cal G}' \hspace{-1mm} = \hspace{-1mm} \left( {{\cal K},{\cal E}'} \right)$}, where {\small ${\cal E}' = \left\{ {{{\tilde e}_{k,j}}= {D_{{\rm{EMD}}}}\left( {k,j} \right)|{e_{k,j}} \ge {e_{{\rm{th}}}},{v_{k,j}} \ge {v_{{\rm{th}}}}} \right\}$}.

In order to maximize clustered data sharing gains, we devise the Distribution-based Adaptive Clustering Algorithm (DACA) to address the privacy-preserving and communication-efficient clustering problem. Our approach employs Theorem \ref{thm} to transform the problem $\mathcal{P}1'$ into a portable operation, which involves selecting the least cluster heads with low ${D_{{\rm{EMD}}}}$ to cover the whole constrained graph ${\cal G}' \hspace{-1mm} = \hspace{-1mm} \left( {{\cal K},{\cal E}'} \right)$. This transformation can be easily explained: if the clusters are not the least, it means that a node with a higher ${D_{{\rm{EMD}}}}$ is selected as the cluster head, contradicting condition 3. As outlined in Algorithm \ref{algorithm_DACA}, the procedure consists of two steps: cluster heads selection and cluster members association. To select cluster heads, we calculate all ${D_{{\rm{EMD}}}}\left( k \right)$ and sort them in descending order (Line 6), and then select cluster heads until all nodes are covered (Line 9). For cluster members, each node is allowed to associate with the best head within the constrained graph ${\cal G}'$ (Line 11).

Our proposed DACA offers several advantages over existing approaches. This algorithm addresses the complex clustering problem in a communication-efficient and privacy-preserving manner. It does not rely on the clustering number as the priori parameter, making it adaptive to different constrained thresholds. Moreover, the proposed DACA achieves optimal cluster formation with low computational complexity. We demonstrate the effectiveness of our approach through simulations and show that it outperforms existing methods in Section \ref{Experiment_Result}. It is worth noting that this approach provides a practical solution for similar constraints-confined clustering problems.

\begin{algorithm}[t]
\small{
\caption{Distribution-based Adaptive Clustering Algorithm (DACA)}\label{algorithm_DACA}
\KwIn{the social closeness graph ${{\cal G}} = \left( {{{\cal K}},{{\cal E}}} \right)$; \\
the constraints ${e_{{\rm{th}}}}$, ${v_{{\rm{th}}}}$.
}
\KwOut{${\cal M}$, ${\cal C}$}
\For{$k \in \mathcal{K}$}{
\For{$j \in \mathcal{K}\backslash \left\{ k \right\}$}{
Compute ${v_{k,j}}$ by ${v_{k,j}} = B_m^s{\log _2}\left( {1 + \gamma \left( {{d_{k,j}}} \right)} \right)$\;
Compute ${D_{{\rm{EMD}}}}\left( {k,j} \right)$ by (\ref{Eq:condition_3});
}
Compute ${D_{{\rm{EMD}}}}\left( k \right)$ by (\ref{Eq:define_EMD}) and sort it\;
}
Build the connected graph ${\cal G}' = \left( {{\cal K},{\cal E}'} \right)$, where {\small ${\cal E}' = \left\{ {{{\tilde e}_{k,j}}= {D_{{\rm{EMD}}}}\left( {k,j} \right)|{e_{k,j}} \ge {e_{{\rm{th}}}},{v_{k,j}} \ge {v_{{\rm{th}}}}} \right\}$ }\;
Select $\cal M$ in descending order ${D_{{\rm{EMD}}}}\left( k \right)$, making all nodes can be connected by cluster heads $\cal M$\;
\While{${{\cal M}} \cup {{\cal C}} \ne {{\cal K}}$}
{Choose $m \in {\cal M},c \in {{\cal K}}\backslash \left( {{{\cal M}} \cup {{\cal C}}} \right)$ with $\max \, {\tilde e_{m,c}}$\;
${{{\cal C}}_m} \leftarrow {{{\cal C}}_m} \cup \left\{ c \right\}$\;
${{\cal C}} \leftarrow \left\{ {...,{{{\cal C}}_m},...} \right\}$\;}
}
\end{algorithm}
\vspace{-2mm}

\subsection{Joint Computed Frequency and Shared Data Volume Optimization Algorithm}
After determining the optimal clustering sets ${\cal{M}}^*$ and ${\cal{C}^*}$ using Algorithm \ref{algorithm_DACA}, we proceed to jointly optimize computed frequency $\boldsymbol{F}$ and shared data volume $\boldsymbol{{N}^s}$. To enhance the clarity of the relationship between variables and function, $\mathcal{P}2$ can be reformulated as follows:
\addtocounter{equation}{0}
\begin{equation}\label{eq:Sub_problem2.2}
\begin{split}
{\mathcal{P}2':}\;\; \mathop {\min }\limits_{\boldsymbol{F},{{\boldsymbol{N}}^s}} \;\;\; {\tau ^s}\left( {\boldsymbol{N}}^s \right) +T\left( {{\boldsymbol{N}}^s} \right)\tau \left( {{{\boldsymbol{N}}^s},\boldsymbol{F}} \right),\quad\quad
\end{split}
\end{equation}
\vspace{-0.8cm}
\begin{align}
&\!\!\!\!\!\!\!\!\!\! \rm{s.\;t.} \scalebox{1}{$\;\;\;\;  0\le n_m^s \le {n_m}, \;\; \forall m \in \mathcal{M},$} \tag{\theequation{a}} \label{C_3_a}\\
&\scalebox{1}{$\;\;\;\; 0 \le {f_k} \le {f_{\max }}, \;\; \forall k \in \mathcal{K}, $} \tag{\theequation{b}} \label{C_3_b}\\
&\scalebox{1}{$\;\;\;\;  {\gamma _k} \le {\gamma _{{\text{th}}}},\;\;\forall k \in \mathcal{K}.$} \tag{\theequation{c}} \label{C_3_c}
\end{align}
This is a joint optimization problem that can be decoupled and solved iteratively. To optimize $\boldsymbol{F}$, we can rewrite the objective function given by (\ref{eq:Sub_problem2.2}) as follows:
\begin{equation}\label{Eq:Sub_P2_objectF}
	{\tau \left( \boldsymbol{F} \right) = \mathop {\max }\limits_{k \in \mathcal{K}} \left\{ {\tau _k^u + \tau _k^{cop}\left( {{f_k}} \right)} \right\}}.
\end{equation}
It is evident that the function $\tau \left( f_k \right)$ is a decreasing function with respect to $f_k$. Therefore, given the clustered shared data volume $n_m^s$, the optimal transmit power of each user $k$, $f_k$ is given by
\begin{equation}\label{optimal_solution_F}
	{{f_k^*} = \min \left\{ {{f_{\max }},{f_{k,\gamma }}} \right\}},
\end{equation}
where ${f_{k,\gamma }}$ satisfies the equality $\gamma _k^u + \varsigma {L_k}E{\tilde n_k}f_{k,\gamma }^2 = {\gamma _{th}}$.
When optimizing the shared data volume $\boldsymbol{{N}^s}$, the objective of subproblem $\mathcal{P}2$ can be reformulated as follows:
\addtocounter{equation}{0}
\begin{equation}\label{eq:Sub_problem2.3}
\begin{split}
{\mathcal{P}3:}\;\; \mathop {\min }\limits_{\boldsymbol{N}^s} \;\;\; {\tau ^s}\left( {\boldsymbol{N}^s} \right) + T\left( {\boldsymbol{N}^s} \right)\tau \left( {\boldsymbol{N}^s} \right),\quad\quad
\end{split}
\end{equation}
\vspace{-0.8cm}
\begin{align}
&\!\!\!\!\!\!\!\!\!\! \rm{s.\;t.} \scalebox{1}{$\;\;\;\; 0\le n_m^s \le {n_m}, \;\; \forall m \in \mathcal{M}, $} \tag{\theequation{a}} \label{C_2-3_a}\\
&\scalebox{1}{$\;\;\;\;  {\gamma _k} \le {\gamma _{{\text{th}}}},\;\;\forall k \in \mathcal{K}.$} \tag{\theequation{b}} \label{C_2-3_b}
\end{align}

To address the no closed-form challenge for $T\left( \cdot \right)$, we employ data fitting to estimate the uncertain function parameters. Due to the complexity involved in directly fitting the function that incorporates multiple variables ${\boldsymbol{N}^s}$, we instead utilize an intermediate function ${\tilde D_{{\text{EMD}}}}\left( {\boldsymbol{N}^s} \right)$ to capture the relationship between $T$ and data quality. Specifically, the number of communication rounds with respect to the data heterogeneity can be expressed as\footnote{The detailed proof process can be refered to Appendix B, Theorem 2 or online version \cite{Faster_FL_arxiv}, and similar results also appear in \cite{FL_with_noniid_in_WN}.}
\begin{equation}\label{data_fitting_T_EMD}
	{T\big( {{{\tilde D}_{{\text{EMD}}}}} \big) = \frac{1}{{{\beta _1}\tilde D_{{\text{EMD}}}^2 + {\beta _2}{{\tilde D}_{{\text{EMD}}}} + {\beta _3}}}}.
\end{equation}
The values of ${\boldsymbol{\beta }} = \left\{ {{\beta _{1}},{\beta _{2}},{\beta _3}} \right\}$ can be estimated by using a sampling set of experimental training result. Considering the uncertainty of the estimated parameters $\boldsymbol{\beta }$, the subproblem $\mathcal{P}3$ can be expressed as a stochastic non-convex problem:
\addtocounter{equation}{0}
\begin{equation}\label{eq:Sub_problem2.4}
\begin{split}
{\mathcal{P}3':}\mathop {\min }\limits_{{{\boldsymbol{N}}^s}}  \Psi \left( {\boldsymbol{\beta} ,{{\boldsymbol{N}}^s}} \right) \hspace{-1mm} \buildrel \Delta \over = \hspace{-1mm}{\mathbb{E}_{\boldsymbol{\beta}} }\big[ {{\tau ^s} \hspace{-1mm}\left( {{{\boldsymbol{N}}^s}} \right) \hspace{-1mm} +  \hspace{-1mm}T\big( {{{\tilde D}_{{\text{EMD}}}}} \big)\tau \left( {{{\boldsymbol{N}}^s}} \right)} \big],\\
\rm{s.\;t.} \scalebox{1}{\;\;\;\;\;\;(\ref{C_2-3_a}) and (\ref{C_2-3_b}).}\quad\quad\quad\quad\quad\quad\quad
\end{split}
\end{equation}

\begin{algorithm}[t]
\small{
\caption{Joint computed Frequency and shared data Volume Optimization (JFVO) Algorithm}\label{algorithm_Sub2}
\KwIn{${{\boldsymbol{N}}_0^s}$ (initial estimate); $I$ number of inner iterations; $J$ number of outer iterations;
${\tilde w_i}$ weights in $\left( {0,\left. 1 \right]} \right.$.
}
\KwOut{${{\boldsymbol{N}}^s}$, ${\boldsymbol{F}}$.}
\For(\tcp*[f]{Outer iteration}){$j=0,...,J$}{
Update ${f_{j,k}}$ based on Eq. (\ref{optimal_solution_F}), for all $k \in \mathcal{K} $\;
Initialize the approximate surrogate: ${\bar g_0}:{{\boldsymbol{N}}^s} \mapsto \frac{\rho }{2}\left\| {{{\boldsymbol{N}}^s} - {\boldsymbol{N}}_{0,j}^s} \right\|$\;

\For(\tcp*[f]{Inner iteration Stochastic SCA Algorithm}){$i=0,...,I$}
{Draw a training point ${{\boldsymbol{\beta }}_{i,j}}$\;
Choose a surrogate function ${g_{i,j}}:{{\boldsymbol{N}}^s} \mapsto {\Psi}\left( {{{\boldsymbol{\beta }}_{i,j}},{{\boldsymbol{N}}^s};{{\boldsymbol{F}}_j}} \right)$ near $\boldsymbol{N}_{i,j}^s$\;
Update the approximate surrogate ${\bar g_{i,j}} = \left( {1 - {\rho _i}} \right){\bar g_{i - 1,j}} + {\rho _i}{g_{i,j}}$\;
Solve the following problem to obtain ${\bar{\boldsymbol{N}}}_{i,j}^s$: ${\bar{\boldsymbol{N}}}_{i,j}^s = \mathop {\arg \min }\limits_{{{\boldsymbol{N}}^s}} {\bar g_{i,j}}\left( {{{\boldsymbol{N}}^s}} \right)$\\
$\quad\quad\quad\quad\;\rm{s.\;t.}\;\;\;\;(\ref{C_2-3_a}) \; and \; (\ref{C_2-3_b})$\;
Update ${\boldsymbol{N}}_{i + 1,j}^s = \left( {1 - {\mu _i}} \right){\boldsymbol{N}}_{i,j}^s + {\mu _i}\overline {\boldsymbol{N}} _{i,j}^s$\;
}
${\boldsymbol{N}}_{0,j+1}^s \leftarrow \boldsymbol{N}_{i+1,j}^s$\;
}
}
\end{algorithm}

It is known that $\Psi \left( {\boldsymbol{\beta} ,{{\boldsymbol{N}}^s}} \right)$ is non-convex and involves the expectation over the random sate $\boldsymbol{\beta}$. The closed-from solution of $\mathcal{P}3$ is difficult to obtain, so we utilize a stochastic successive convex approximation (SSCA) algorithm \cite{SSCA} to construct a recursive convex approximation of subproblem $\mathcal{P}3$.
As illustrated in Algorithm \ref{algorithm_Sub2}, a joint optimization method is proposed by optimizing $\boldsymbol{F}$ and ${\boldsymbol{N}}^s$ iteratively. At the $j$-th outer iteration, the computed frequency of each user $f_{j,k}$ can be updated by (\ref{optimal_solution_F}). It is clear that ${\tau ^s} \hspace{-1mm}\left( {{{\boldsymbol{N}}^s}} \right)$ is a convex function, but $T\big( {{{\tilde D}_{{\text{EMD}}}}} \big)\tau \left( {{{\boldsymbol{N}}^s}} \right)$ the right term of $\mathcal{P}3$ is non-convex. So at the $i$-th inner iteration, we design a recursive convex approximation ${\bar g_{i,j}}$ of the original objective function, which is updated by
\begin{equation}\label{Recursive_surrogate_function_}
{{\bar g_{i,j}} = \left( {1 - {\rho _i}} \right){\bar g_{i - 1,j}} + {\rho _i}{g_{i,j}},}
\end{equation}
where $\rho _i$ satisfies ${\rho _i} \to 0,\sum\limits_{i \to \infty } {{\rho _i} = } \infty ,\sum\limits_{i \to \infty } {{\rho _i}^2}  < \infty$. $g_{i,j}$ is a convex approximation of $\Psi \left( {\boldsymbol{\beta}_{i,j} ,{{\boldsymbol{N}}^s}} \right)$ at inter iteration $i$, which is given by
\begin{equation}\label{Surrogate_function_g}
\begin{split}
{g_{i,j}} &= {\tau ^s}\left( {\boldsymbol{N}^s} \right) + \tilde \Psi \left( {\boldsymbol{N}_{i - 1,j}^s} \right) + \nabla \tilde \Psi \left( {\boldsymbol{N}_{i - 1,j}^s} \right) \\
& * \big( \boldsymbol{N}^s - N_{i - 1,j}^s \big) + \frac{L}{2}{\left\| {\boldsymbol{N}^s - \boldsymbol{N}_{i - 1,j}^s} \right\|^2},
\end{split}
\end{equation}
where $\tilde \Psi \!=\! T({\tilde D_{{\text{EMD}}}})\tau \left( {{{\boldsymbol{N}}^s}} \right)$ and $L$ is Lipschitz constant. Then we obtain the optimal solution ${\bar{\boldsymbol{N}}}_{i,j}^s$ by solving the following problem:
\addtocounter{equation}{0}
\begin{equation}\label{eq:Sub_problem_S_avg}
\begin{split}
{\bar{\boldsymbol{N}}}_{i,j}^s = \mathop {\arg \min }\limits_{{{\boldsymbol{N}}^s}} {\bar g_{i,j}}\left( {{{\boldsymbol{N}}^s}} \right),\\
\rm{s.\;t.} \scalebox{1}{\;\;\;\;\;\;(\ref{C_2-3_a}) and (\ref{C_2-3_b})}.\quad\quad\quad
\end{split}
\end{equation}

Obviously, this is a convex problem which can be effectively solved by off-the-shelf solvers such as CVX. After that, ${\boldsymbol{N}}^s$ is updated according to
\addtocounter{equation}{0}
\begin{equation}\label{eq:Update_Ns}
\begin{split}
{\boldsymbol{N}}_{i + 1,j}^s = \left( {1 - {\mu _i}} \right){\boldsymbol{N}}_{i,j}^s + {\mu _i}\overline {\boldsymbol{N}} _{i,j}^s,
\end{split}
\end{equation}
where $\mu _i$ satisfy ${\mu _i} \to 0,\sum\limits_{i \to \infty } {{\mu _i} = } \infty, \sum\limits_{i \to \infty } {{\mu _i}^2}  < \infty$ and $\mathop {\lim }\limits_{i \to \infty } {\mu _i}/{\rho _i} = 0$.

\begin{figure}[t]
\begin{center}
\includegraphics[width=8.5cm,height=6.2cm]{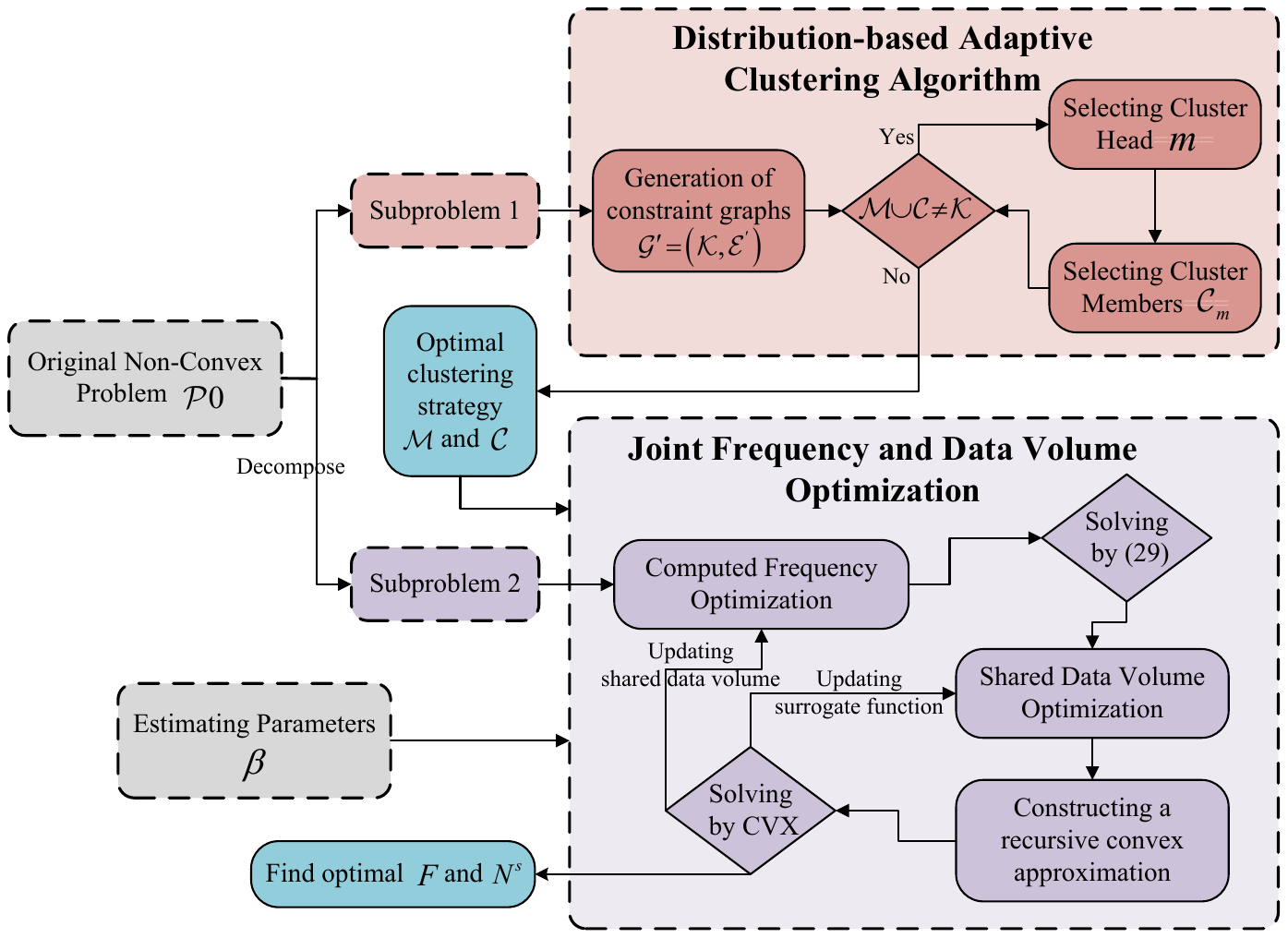}
\caption{Flow chard of the proposed algorithms.}
\vspace{-3mm}
\label{fig_algorithm}
\vspace{-3mm}
\end{center}
\end{figure}

In summary, the flowchart for addressing the original non-convex problem is illustrated in Fig. \ref{fig_algorithm}. This process decomposes into two main subproblems: the selection of cluster heads and members, addressed efficiently by Algorithm \ref{algorithm_DACA}; and the joint optimization of computed frequency and shared data volume. The latter subproblem is tackled by Algorithm \ref{algorithm_Sub2}, which involves the outer iteration to solve for the optimal computed frequency in (\ref{optimal_solution_F}) and the inner iteration to update the shared data volume in (\ref{eq:Update_Ns}).
%
%
%By setting $\tilde \Psi \!=\! T({\tilde D_{{\text{EMD}}}})\tau \left( {{{\boldsymbol{N}}^s}} \right)$, we construct a $L$-lipschitz gradient surrogate function for $\mathcal{P}3$, which is given by
%\begin{equation}\label{Surrogate_function_g}
%\begin{split}
%{g_{i,j}} &= {\tau ^s}\left( {\boldsymbol{N}^s} \right) + \tilde \Psi \left( {\boldsymbol{N}_{i - 1,j}^s} \right) + \nabla \tilde \Psi \left( {\boldsymbol{N}_{i - 1,j}^s} \right) \\
%& * \big( \boldsymbol{N}^s - N_{i - 1,j}^s \big) + \frac{L}{2}{\left\| {\boldsymbol{N}^s - \boldsymbol{N}_{i - 1,j}^s} \right\|^2}
%\end{split}
%\end{equation}
%
%We construct a L-lipschitz gradient surrogate function for $\tilde \Psi \!=\! T({\tilde D_{{\text{EMD}}}})\tau \left( {{{\boldsymbol{N}}^s}} \right)$, which is given by
%\begin{equation}\label{Surrogate_function_g}
%\begin{split}
%{g_{i,j}} &= {\tau ^s}\left( {\boldsymbol{N}^s} \right) + {\tilde \Psi _{i,j}} + \nabla {\tilde \Psi _{i,j}}\left( {\boldsymbol{N}^s - {\boldsymbol{N}_{i - 1,j}^s}} \right) \\
%& + \frac{L}{2}{\left\| {\boldsymbol{N}^s - {\boldsymbol{N}_{i - 1,j}^s}} \right\|^2}
%\end{split}
%\end{equation}
%where $\tilde \Psi _{i,j}$ is the

\subsection{Complexity Analysis}
Without loss of generality, we discuss the computational complexity of the proposed DACA and JFVO algorithms. When addressing the clustering problem, the predominant computational consumption arises from the generation of the constraint graph. The complexity of calculating these constraint edges in DACA is ${\rm O}\left( {{K^2}} \right)$. For the joint optimization problem, the computational complexity mainly relies on the calculation of the computed frequency, the construction of a recursive convex approximation function, and the solution of the shared data volume. In the outer iteration, the complexity of updating frequency $\boldsymbol{F}$ is ${\rm O}\left( {{JK}} \right)$. During the inner iteration, we design a recursive convex approximation for optimizing the shared data volume. The complexity in calculating the surrogate function is ${\rm O}\left( {{JIM}} \right)$, where $M$ is the cluster number determined by DACA. Problem (35) is a quadratic problem (QP) \cite{NumericalOptimization} and the complexity of solving it using a standard convex optimization method is denoted by ${\rm O}\left( {{JIM^2}} \right)$.

In summary, besides the advantages of the DACA discussed in Section \ref{Sec_DACA}, our proposed algorithm can effectively accomplish users group and accelerate FEL training in uncertain environments. Stochastic optimization algorithms can mitigate the instability resulting from parameter estimation. Moreover, the computational requirement of DACA and JFVO is affordable, making it suitable for practical use as an online algorithm.
\vspace{-3mm}

\section{Simulation and Experimental Results}\label{Sec_SimuRe}
In this section, we present the results of both numerical simulations and experiments to evaluate the performance of our proposed framework. We employ federated learning for image classification using a circular network area with a radius $r=1,000m$, and the specific parameters used in the simulations are comprehensively enumerated in Table \ref{table_1}.

\begin{table}[t]
\centering
\vspace*{-0.3cm}
\caption{\vspace*{-0.05em}SYSTEM PARAMETERS}\vspace*{-0.6em}
\label{table_1}
\resizebox{8.3cm}{!}{
\renewcommand \arraystretch{1.3}
\begin{tabular}{ c  c }
    \toprule
    \\[-5mm]
    \textbf{Parameter}         & \textbf{Value}               \\
    \hline
    \hline
    {Multicast interference} $I$                       & 3 dB              \\
     %$f_c^s$                   & 28 GHz                 &  $\gamma_{\text{th}}$            & 0.005 J                  \\

     %$\beta_{\text{PL}}$       & 32.4 dB                &  $\zeta$                         & 2.1                 \\

     %$M_{I}$                   & 3 dB                   &  $M_{S}$                         & 4 dB                \\

     {Number of subcarriers} $R$                       & 10          \\
     {Nosie power spectral density}  $N_0$                           & -174 dBm/Hz              \\

     {Transmit power of the BS} $P_{\text{BS}}$           & 1 W             \\
     {Transmit power of user} $k$ $P_k$                           & 0.01 W                   \\

     {Bandwidth of multicasting} $B_m^s$                      & 1 GHz        \\
     {Bandwidth of broadcating} {$B^d$}                    & 20 MHz         \\
     {Bandwidth of uploading} $ B^u $                    & 1 MHz           \\
     {Energy consumption requirement} $\gamma_{\text{th}}$            & 0.005 J                  \\
     {Energy consumption coefficient} $\varsigma $                    & $4 \times 10^{-26}$      \\
     {Maximum computation frequency}  $f_{\max }$                     & $1.2 \times 10^3$ MHz    \\

     {CPU cycles for per sample} $L_k$                           & $2.5 \times 10^5$ cycles/sample         \\
    \bottomrule
\end{tabular}
}
\vspace*{-0.5cm}
\end{table}

%\begin{table}[t]
%\centering
%\vspace*{-0.3cm}
%\caption{\vspace*{-0.05em}SYSTEM PARAMETERS}\vspace*{-0.6em}
%\label{table_1}
%\resizebox{8.3cm}{!}{
%\renewcommand \arraystretch{1.3}
%\begin{tabular}{ c  c  c  c }
%    \toprule
%    \\[-5mm]
%    \textbf{Parameter}         & \textbf{Value}         & \textbf{Parameter}               & \textbf{Value}           \\
%    \hline
%     Multicast interference $I$                       & 3 dB                 &  Energy consumption requirement $\gamma_{\text{th}}$            & 0.005 J                  \\
%     %$f_c^s$                   & 28 GHz                 &  $\gamma_{\text{th}}$            & 0.005 J                  \\
%
%     %$\beta_{\text{PL}}$       & 32.4 dB                &  $\zeta$                         & 2.1                 \\
%
%     %$M_{I}$                   & 3 dB                   &  $M_{S}$                         & 4 dB                \\
%
%     $R$                       & 10                     &  $N_0$                           & -174 dBm/Hz              \\
%
%     $P_{\text{BS}}$           & 1 W                    &  $P_k$                           & 0.01 W                   \\
%
%    {$B^d$}                    & 20 MHz                 &  $f_{\max }$                     & $1.2 \times 10^3$ MHz    \\
%
%    $ B^u $                    & 1 MHz                  &  $L_k$                           & $2.5 \times 10^5$ cycles/sample         \\
%
%    $B_m^s$                      & 1 GHz                  &  $\varsigma $                    & $4 \times 10^{-26}$      \\
%    \bottomrule
%\end{tabular}
%}
%\vspace*{-0.5cm}
%\end{table}

\subsection{Experimental Setup}
\emph{1) Heterogeneous Data Setting:}

\emph{Datasets:} We evaluate our results using the MNIST \cite{MNIST} , CIFAR-10 datasets \cite{CIFAR10} and Shakespeare \cite{LEAF}, which are commonly used image classification and next word prediction.  The MNIST dataset consists of a training dataset with 60,000 images and a test dataset with 10,000 images. The CIFAR-10 dataset consists of 10 classes of colored objects with 50,000 training and 10,000 test samples.  {Shakespeare contains text dataset from the complete works of William Shakespeare and dialogue of each character is used as one dataset on a device. Naturally, Shakespeare dataset exhibits feature distribution skew.}
The processing details for these non-IID datasets are as follows:

\emph{The label distribution skew:} To create disjoint non-IID client training data, partial users receive data partitions from only a single class, while the remaining users select samples without replacement from the training dataset with all labels.

\emph{The feature distribution skew:} is set by the noise-based feature imbalance \cite{Survey_FL_noniid_Experimental_study}.
The entire dataset is randomly and equally divided among multiple parties. Then all parts add different levels of Gaussian noise to their local datasets to achieve distinct feature distributions. {Shakespeare dataset on devices already have different feature distributions and does not need to be preprocessed. }

\emph{2) Model Structure:}
We adopt Convolutional Neural Networks (CNNs) as they are widely used in image classification tasks. For the MNIST dataset, we employ a $2$-layer CNN with two $5 \times 5$ convolutional layers and two fully connected layers. For the more complex CIFAR-10 dataset, we employ the famous CNN architecture known as VGG-16 \cite{Vgg16}. 
{For the Shakespeare dataset, we adapt a Long Short-Term Memory (LSTM) model as the network architecture of NLP task.}

\begin{figure*}
\begin{center}
\subfloat[\footnotesize{Loss on MNIST data}]{\includegraphics[width=4.0cm,height=2.8cm]{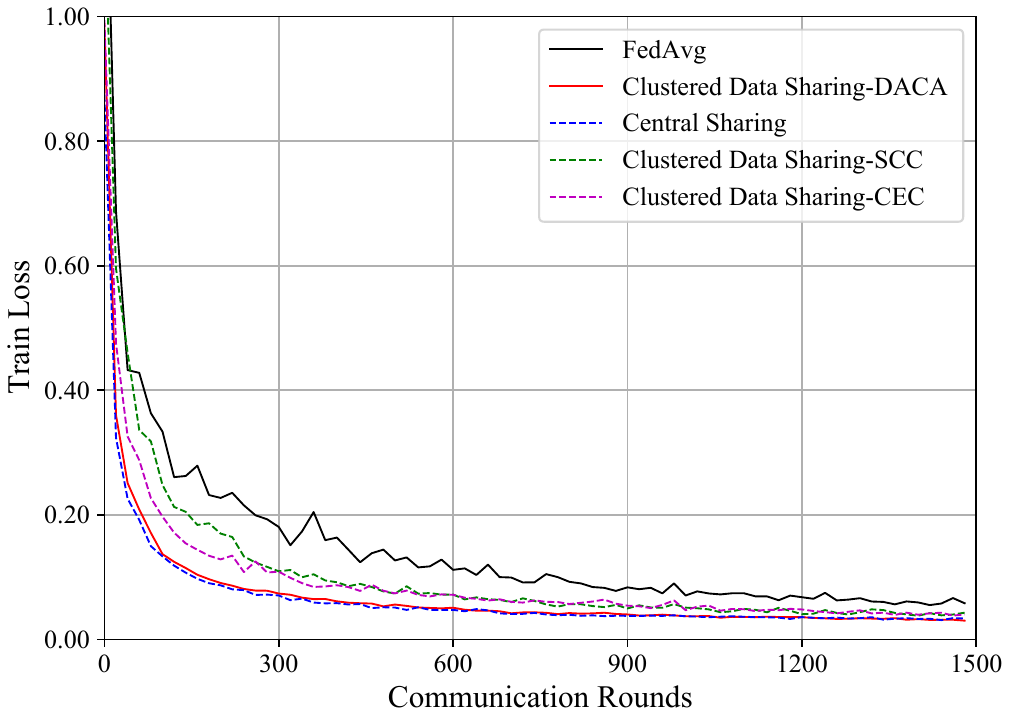}%
\label{MNIST_loss}}
\quad
\subfloat[\footnotesize{Loss on CIFAR-10 data}]{\includegraphics[width=4.0cm,height=2.8cm]{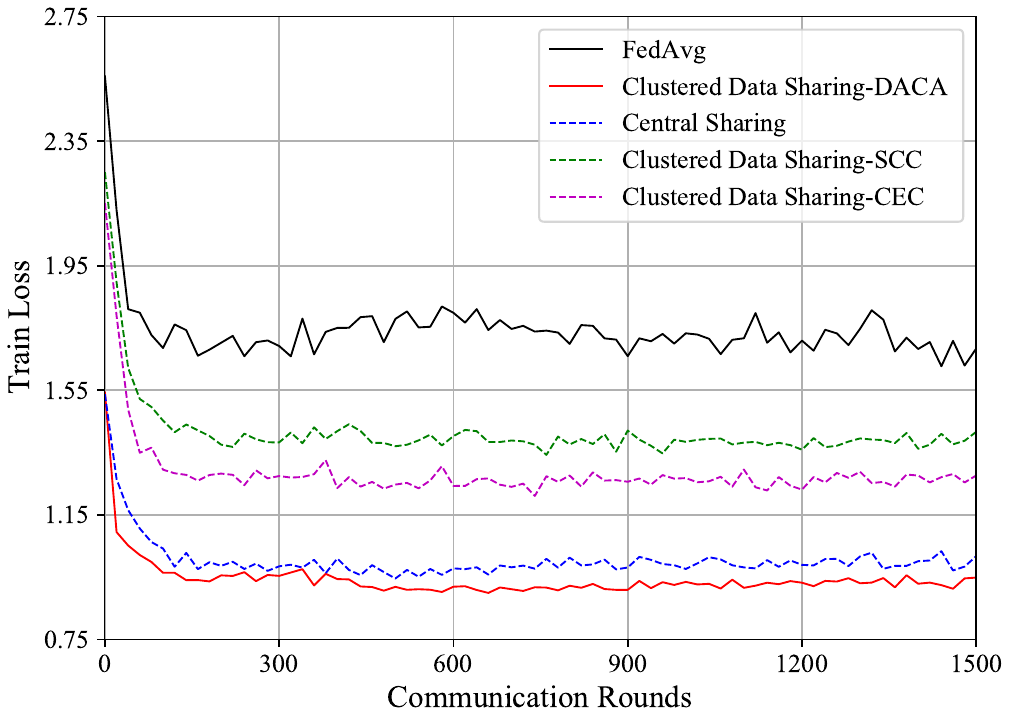}%
\label{CIFAR_Loss}}
\quad
\subfloat[\footnotesize{\hl{Loss on  Shakespeare data}}]
{\includegraphics[width=4.0cm,height=2.8cm]{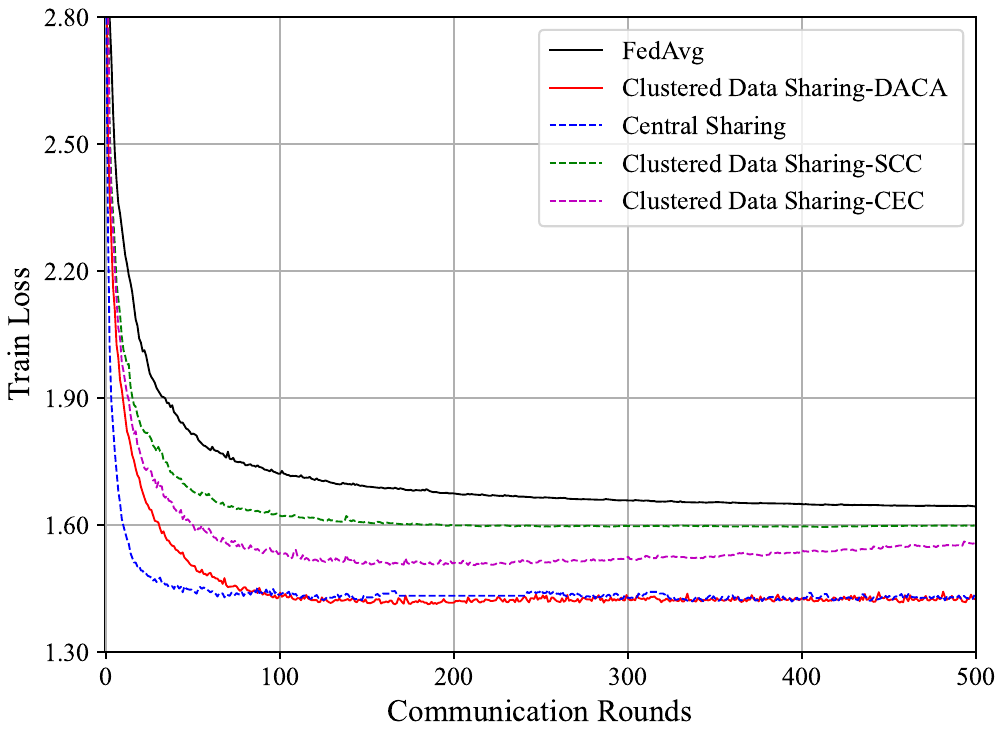}%
\label{ShakeSpeare_Loss}}
\quad
\subfloat[\footnotesize{Loss on Noise-based data}]
{\includegraphics[width=4.0cm,height=2.8cm]{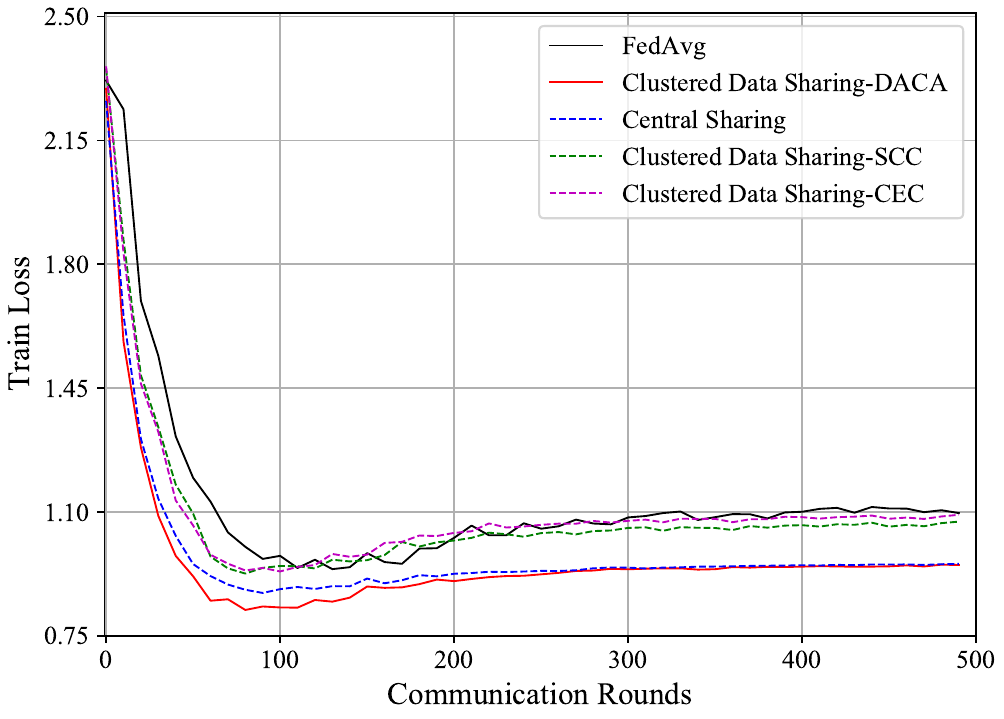}%
\label{Feature_Skewness_Loss}}
\\

\subfloat[\footnotesize{Accuracy on MNIST data}]{\includegraphics[width=4.0cm,height=2.8cm]{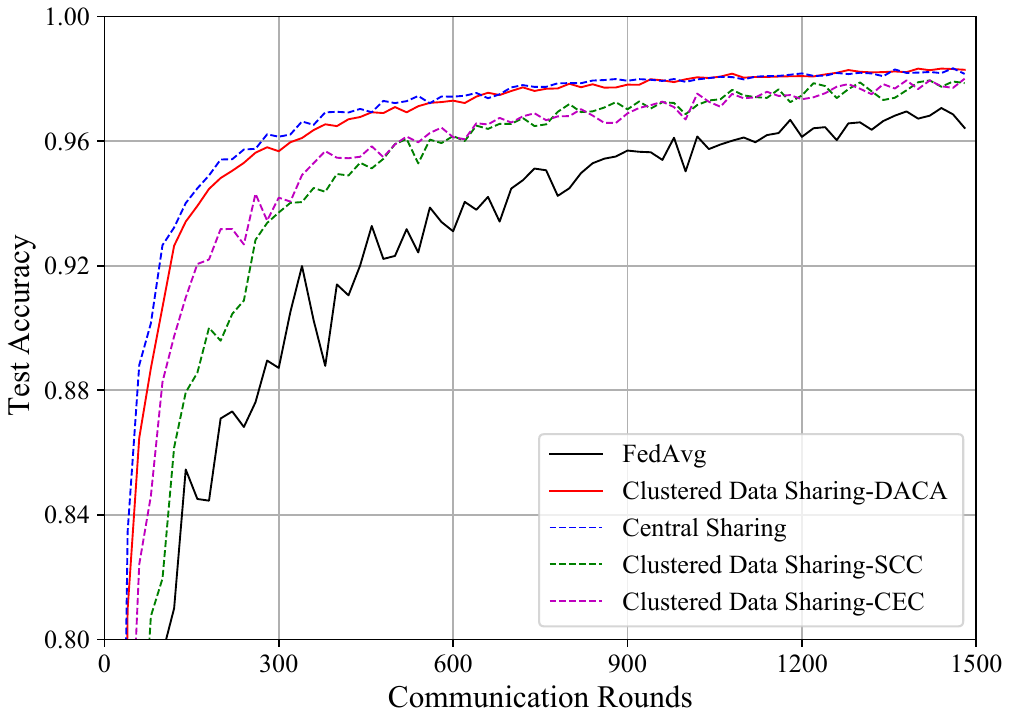}%
\label{MNIST_Acc}}
\quad
\subfloat[\footnotesize{Accuracy on CIFAR-10 data}]{\includegraphics[width=4.0cm,height=2.8cm]{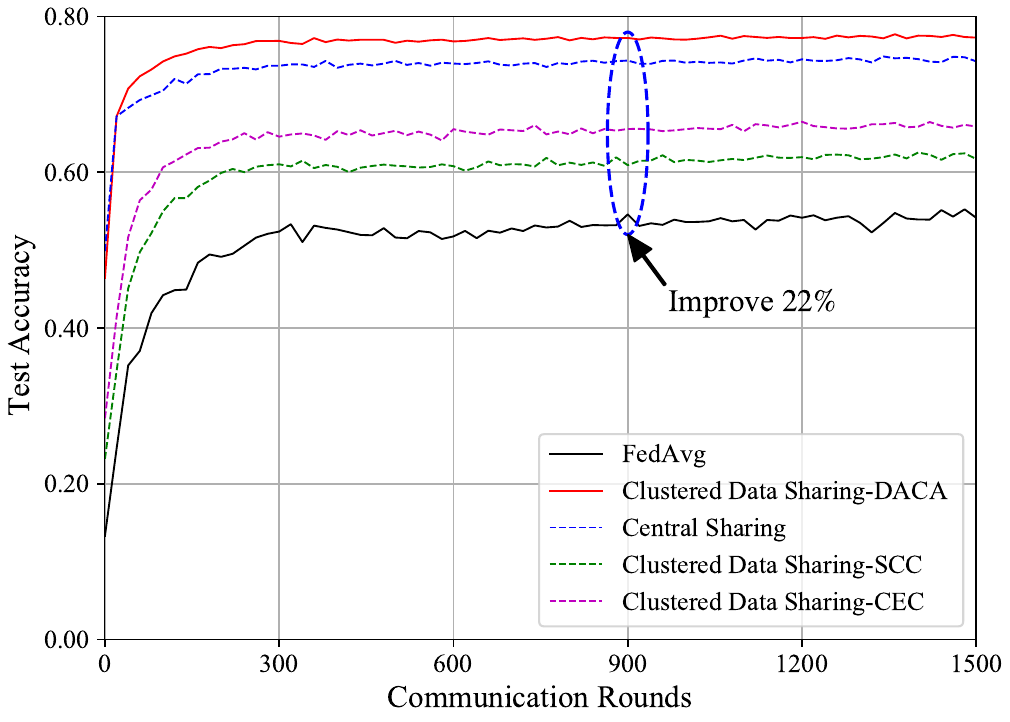}%
\label{CIFAR_Acc}}
\quad
\subfloat[\footnotesize{\hl{Accuracy on Shakespeare data}}]
{\includegraphics[width=4.0cm,height=2.8cm]{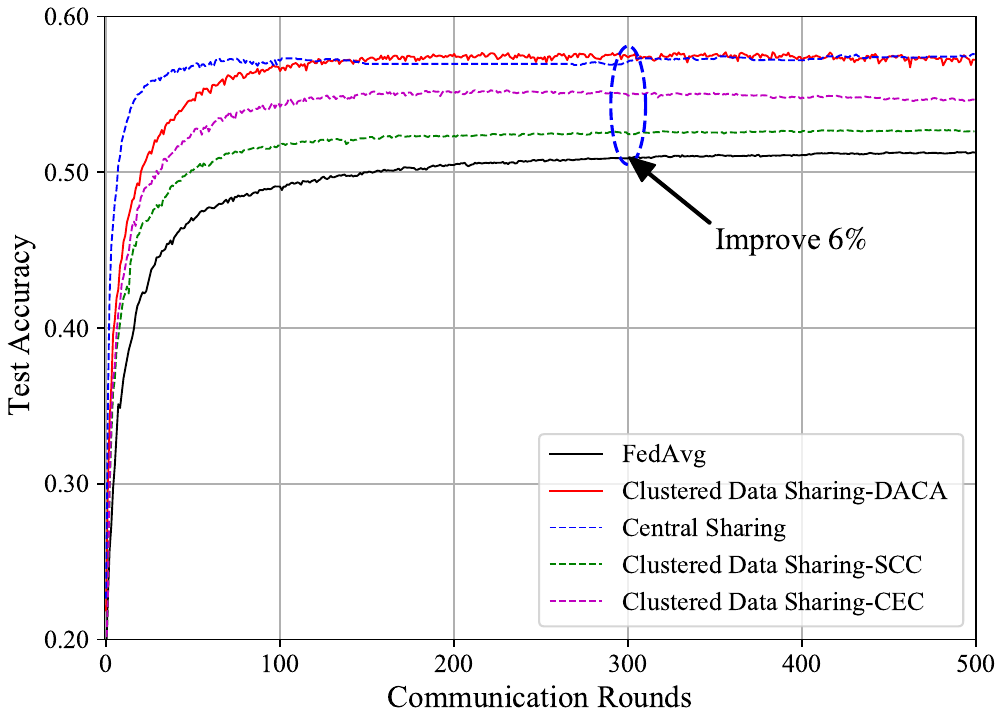}%
\label{ShakeSpeare_Acc}}
\quad
\subfloat[\footnotesize{Accuracy on Noise-based data}]
{\includegraphics[width=4.0cm,height=2.8cm]{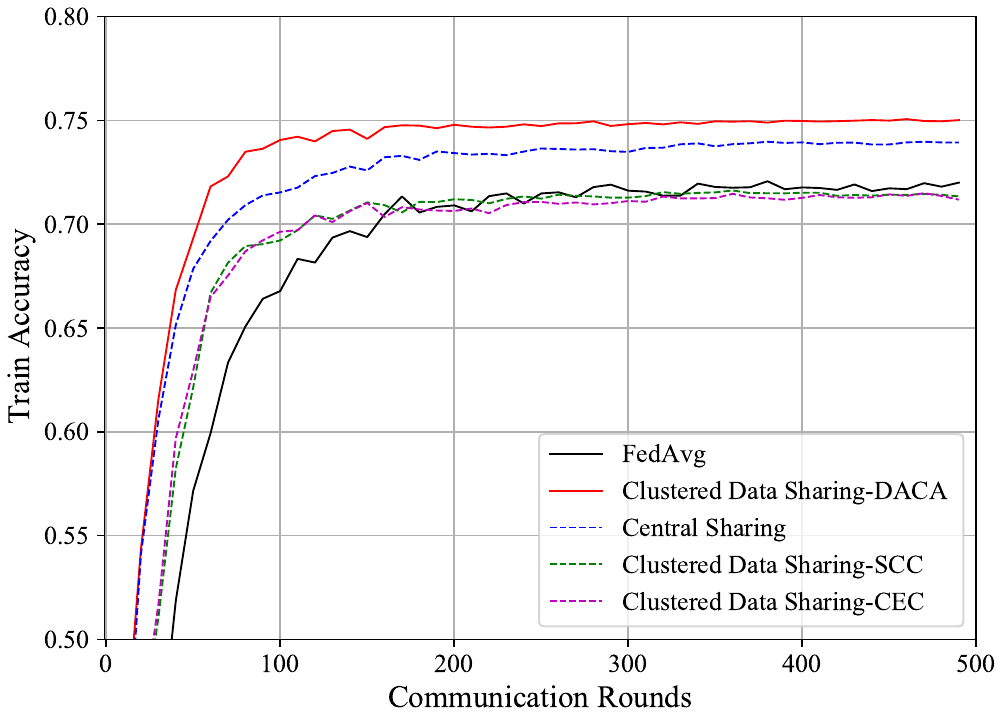}%
\label{Feature_Skewness_Acc}}
\caption{The convergence and accuracy performance with different data sharing strategies, ${v_{th}} = 3.5 \times 10^5 $ bit/s and ${e_{th}} = 0.5$.}
\vspace{-1.5mm}
\label{fig_DACA}
\vspace{-4mm}
\end{center}
\end{figure*}

\begin{figure*}
\begin{center}
\subfloat[\footnotesize{Label distribution of IID data}]{\includegraphics[width=4.9cm,height=3.5cm]{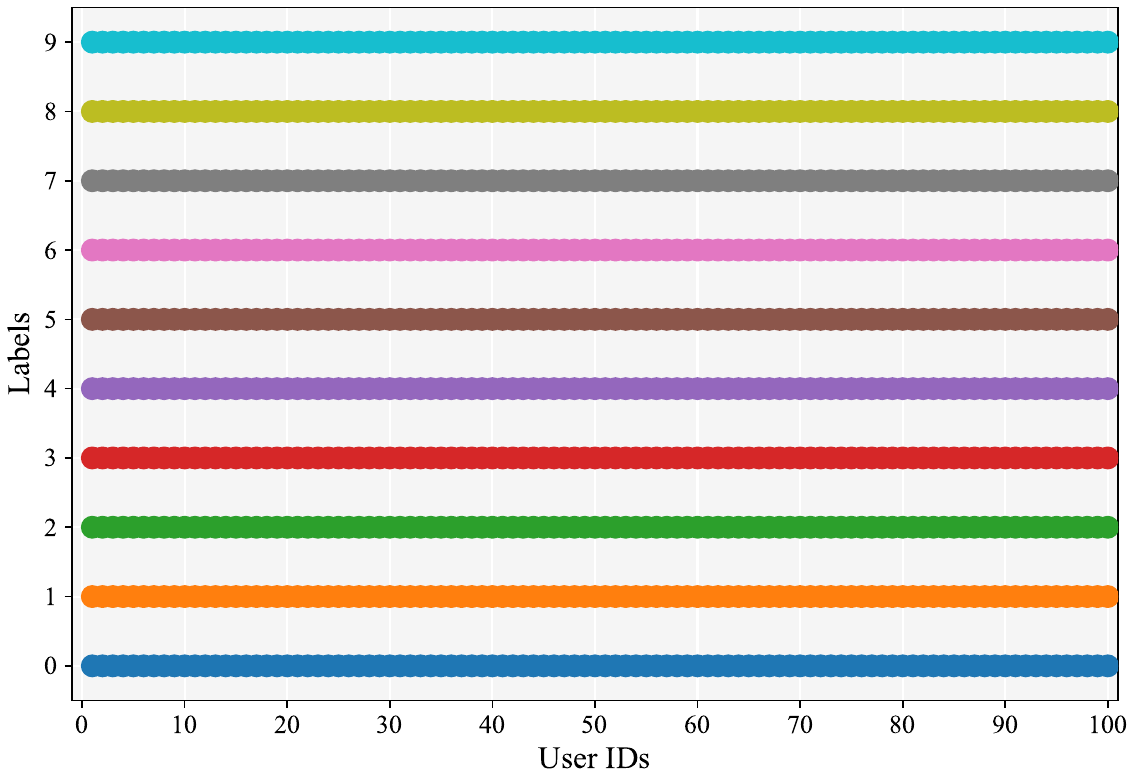}%
\label{fig_iid_distribution}}
\quad\quad
\subfloat[\footnotesize{Label distribution before data sharing}]
{\includegraphics[width=4.9cm,height=3.5cm]{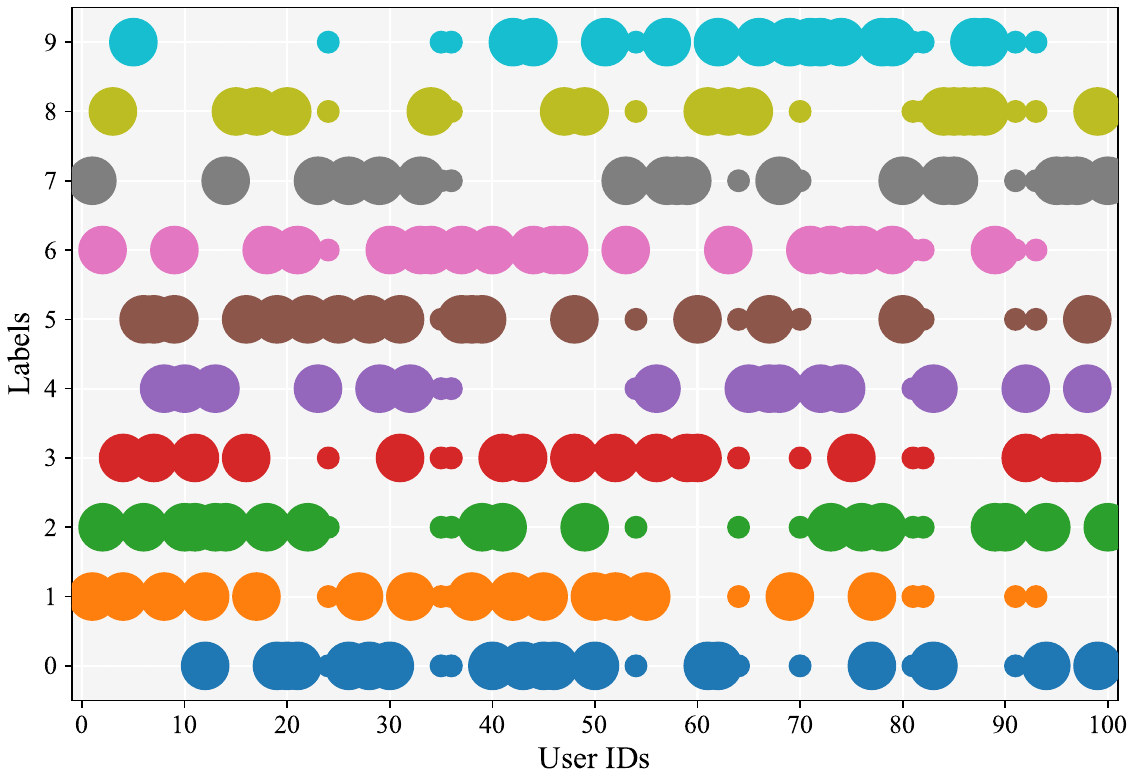}%
\label{fig_distribution_before}}
\quad\quad
\subfloat[\footnotesize{Label distribution after data sharing}]
{\includegraphics[width=4.9cm,height=3.5cm]{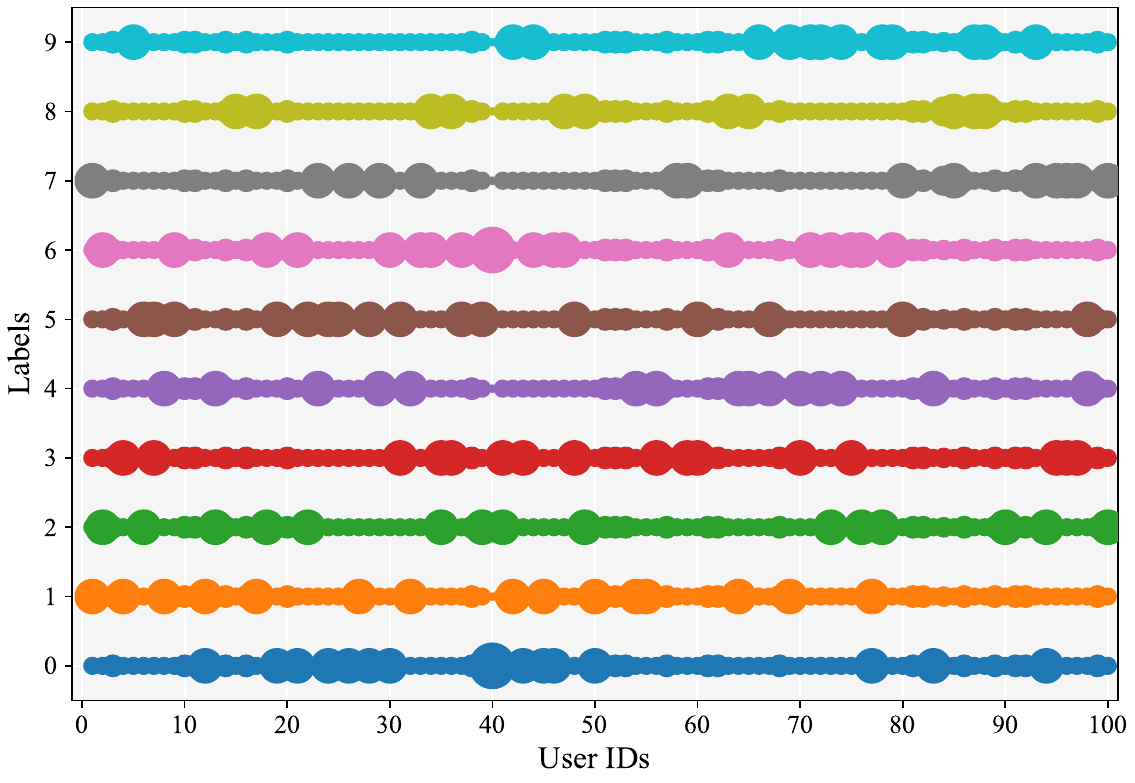}%
\label{fig_distribution_after}}
\caption{The illustration of the sample number per class allocated to each user (indicated by dot sizes).}
\label{fig_distribution}
\vspace{-1.5mm}
\vspace{-4mm}
\end{center}
\end{figure*}

\textbf{Baselines.}
To demonstrate the performance improvement, the clustered data sharing method is compared with two strategies considering maximized social closeness or data volume.
%\addtolength{\topmargin}{0.04in}

\begin{itemize}
\item FedAvg \cite{FedAvg}: Traditional FL with a server and distributed users to train a global model without data sharing.
\item FedAvg with Central Sharing \cite{FL_with_noniid,FL_with_noniid_in_WN}: A portion of the global dataset is sent directly by the server to all users.
\item Social Closeness Clustering (SCC) \cite{SocialClustering}: Clustering the users by social closeness, and the most trusted users are selected as cluster heads.
\item Communication Efficient Clustering (CEC) \cite{D2D_data_sharing_DL_WN}: Clustering the users by transmission rate, and the users with high multicast rates are selected as cluster heads.
\end{itemize}

{To demonstrate the superiority of the resource allocation algorithm, the JFVO algorithm is compared with some existing SOTA  (State-of-the-Art) works:}
\begin{itemize}
\item {The stochastic Majorization-Minimization (MM){\protect~\cite{StochasticMM}}:  This algorithm focuses on constructing a surrogate function that majorizes (upper bounds) the original objective function and iteratively minimizes this surrogate function, ensuring convergence by reducing the original objective at each step.}
\item {Sample Average Approximation (SAA){\protect~\cite{SAA}}: This method can solve stochastic optimization problems by replacing the expectation in objective function with sample average. And the stochastic problem is transformed into a deterministic one that can be solved by standard optimization techniques.}
\end{itemize}

\subsection{Experiment Results} \label{Experiment_Result}

\emph{1)FL Performance of Clustering Strategies:}

In Fig. \ref{fig_DACA}, we compare the proposed DACA with several baselines. Although DACA is primarily designed for label distribution skew, its applicability can be readily extended to other non-IID cases such as feature distribution skew. The left two columns of Fig. \ref{fig_DACA} show the superior performance of the algorithm on label distribution skew, and the right columns shows DACA's generalization in the performance of feature distribution skew. {Moreover, we conduct experiments on the commonly used Dirichlet distribution. The results in Table {\protect\ref{tab:Dir}} confirm that our method remains effective under this setting.}

From the top row of Fig. \ref{fig_DACA}, it is evident that the proposed clustered data sharing methods outperform others in terms of reducing the number of communication rounds and achieving lower loss values. Additionally, the bottom of Fig. \ref{fig_DACA} demonstrate that the proposed framework significantly  enhances test accuracy. {The CIFAR-10 dataset, being more complex in feature distribution compared with MNIST dataset, exhibits more pronounced improvements. This demonstrates that our method can achieve greater performance gains for training with initially poor accuracy.
The DACA shows favorable results even compared with central sharing method on CIFAR-10 and noise-based dataset. This is because central sharing only broadcast the same samples to all participants while clustered sharing multicast different samples to augment local dataset.
Moreover,  the DACA algorithm improves the accuracy by 6\% on the Shakespeare dataset, which highlights the applicability of the proposed framework to natural language processing (NLP) tasks.} These observation demonstrate the superiority of DACA over other clustering strategy in terms of estimating statistical heterogeneity. 

It can be observed from Fig. \ref{fig_distribution}(a) that if the data distribution follows IID distribution, the label distribution across users would be uniform. However, as shown in Fig. \ref{fig_distribution}(b), certain users may lack specific labels in non-IID case, leading to distribution skewness.
In Fig. \ref{fig_distribution}(c), the label distribution after multicasting closely approximates an IID distribution.
This is because that our proposed clustered data sharing method effectively addresses skewness by filling missing labels for high EMD users. 

\begin{table}[t]
\centering
\caption{Comparison of performance on Dirichlet distribution.}
\label{tab:Dir}
\setlength{\tabcolsep}{4pt} 
\renewcommand{\arraystretch}{1.2} 
\begin{tabular}{@{}ll|cccc|cc@{}}
\toprule
\textbf{Dataset} & \textbf{Dirichlet} & \textbf{FedAvg} & \textbf{SCC} & \textbf{CEC} & \textbf{Central} & \textbf{DACA(ours)} \\ 
\midrule
\midrule
\multirow{3}{*}{\textbf{\emph{MNIST}}} 
 & Dir=0.1 & 96.17 & 97.42 & 97.50 & 97.84 & \textbf{97.98}$(\uparrow 1.8)$ \\[-0.2ex] 
 & Dir=1.0 & 97.45 & 97.86 & 97.90 & 98.07 & \textbf{98.09}$(\uparrow 0.6)$ \\[-0.2ex] 
 & Dir=10 & 97.80 & 97.90 & 97.94 & 97.97 & \textbf{98.08}$(\uparrow 0.2)$ \\  
\midrule
\midrule
\multirow{3}{*}{\textbf{\emph{CIFAR-10}}} 
 & Dir=0.1 & 67.72 & 74.00 & 74.93 & 75.73 & \textbf{75.90}$(\uparrow 8.2)$ \\  [-0.2ex] 
 & Dir=1.0 & 75.10 & 75.88 & 76.05 & \textbf{76.65} & 76.48$(\uparrow 1.3)$ \\  [-0.2ex] 
 & Dir=10 & 76.70 & 77.93 & 77.95 & \textbf{78.04} & 77.54$(\uparrow 0.8)$ \\    
\bottomrule
\end{tabular}
\end{table}

%In Fig. \ref{fig3}, the experiment results of training loss and test accuracy are provided. From Fig. \ref{fig3-1}, we can see that the proposed clustered data sharing methods can reduce the number of communication rounds from about 1000 to 400 compared to FedAvg, which is close to central sharing but better than other baselines. From Fig. \ref{fig3-2}, we can see that the proposed FL framework can improve the test accuracy from 96.42$\%$ to 98.72$\%$ compared to FedAvg. These gain terms from the fact that the proposed FL framework reduces the degree of non-IID data, and modifies the deviation of model update by SGD.

%\begin{figure*}[ht]
%\begin{center}
%\subfloat[\footnotesize{Communication Rounds Fitting}]{\includegraphics[width=7.5cm,height=6cm]{data fitting.pdf}%
%\label{fig4-1}}
%\quad\quad\quad\quad
%\subfloat[\footnotesize{Accuracy Fitting}]{\includegraphics[width=8.05cm,height=6cm]{data fitting-acc.pdf}%
%\label{fig4-2}}
%\caption{The performance for communication rounds fitting and accuracy Fitting}
%\vspace{-1.5mm}
%\label{fig_4}
%\end{center}
%\end{figure*}
\begin{figure}[t]
\begin{center}
\includegraphics[width=7.3cm,height=5.5cm]{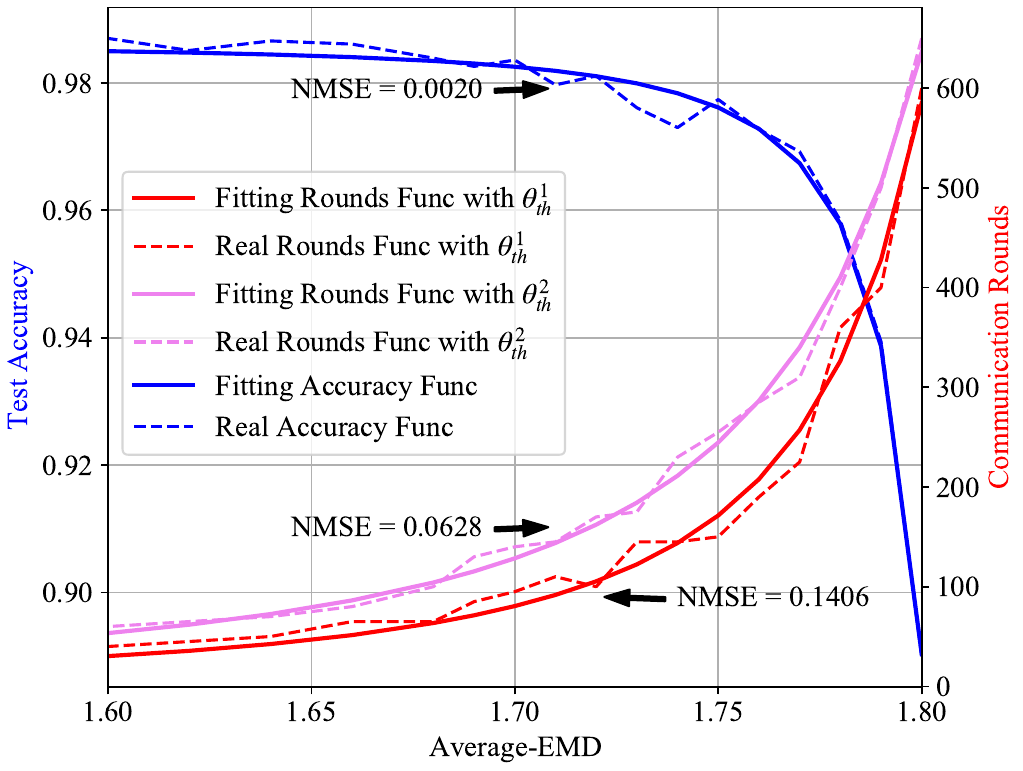}
\vspace{-3mm}
\caption{The performance for communication rounds fitting and accuracy fitting. When the accuracy threshold $\theta_{th}^1$ is set to 0.95, the estimated parameters are calculated to be ${\beta _{1}=0.50}, {\beta _{2}=-1.83}, {\beta _{3}=1.70}$. When the accuracy threshold $\theta_{th}^2$ is set to 0.96, the estimated parameters are calculated to be ${\beta _{1}=0.24}, {\beta _{2}=-0.89}, {\beta _{3}=0.06}$.}
\vspace{-3mm}
\label{fig_fitting}
\end{center}
\end{figure}

\emph{2) Performance for Data Fitting:}

To obtain the communication rounds function $T\left( \cdot \right)$, we conduct a series of FL experiments involving heterogeneous datasets with different average-EMD in Fig. \ref{fig_fitting}. The number of communication rounds is recorded when the test accuracy of the global model first meets the specified threshold. The acquired data from these experiments are used to estimate the fitted parameters $\boldsymbol{\beta}$ in (\ref{data_fitting_T_EMD}). We also record the final accuracy of the global models which are trained on various average-EMD datasets with a sufficient number of rounds. In order to demonstrate the effect of data fitting, the Normal Mean Squared Error (NMSE) serves as an important indicator, which measures the disparities between estimated and actual values.  The predicted values have low NMSE which indicates the high-precision of fitted round-EMD function.

Fig. \ref{fig_fitting} effectively demonstrates the performance of the function $T\left( \cdot \right)$ under different accuracy thresholds. This figure depicts that as the average EMD rises, the number of communication rounds proportionally increases. The rate of increase relatively gradual initially, but then escalates exponentially.  Additionally, as the desired accuracy threshold becomes more demanding, the number of communication rounds correspondingly rises. These observations are consistent with the preliminary experiments in the Section \ref{sec:pre_experiments}, and highlight a trade-off between maximizing global model accuracy and minimizing communication rounds. Furthermore, As shown in the blue line, a discernible relationship exists between accuracy and statistical heterogeneity, and this correlation exhibits an inverse trend in comparison to the communication rounds. Consequently, it is crucial to identify an appropriate average-EMD value for balancing the communication cost and model accuracy, i.e., the optimal shared data volume ${{\boldsymbol{N}}^s}$.

%\begin{table}[h]
%\begin{center}
%\caption{\vspace*{-0.05em}ESTIMATED PARAMETERS FOR FITTING ROUNDS FUNCTION}\vspace*{-0.6em}
%\label{table_2}
%\renewcommand \arraystretch{1.5}
%\begin{tabular}{| c | c | c | c | c |}
%\hline
%\textbf{Accuracy threshold}   & ${\beta _{1}}$          & ${\beta _{2}}$       & ${\beta _{3}}$     & \textbf{NMSE}      \\
%\hline
% $0.95$                       & 0.491                   &  $-1.826$            & 1.697              &  0.1406            \\
%\hline
% $0.96$                       & 0.236                   &  $-0.888$            & 0.836              &  0.0628            \\
%\hline
%\end{tabular}
%\end{center}
%\vspace{-0.5cm}
%\end{table}

\begin{figure}[t]
\begin{center}
\subfloat[\footnotesize{{Convergence versus resource allocation algorithms.}}]{\includegraphics[width=7.3cm,height=5.5cm]{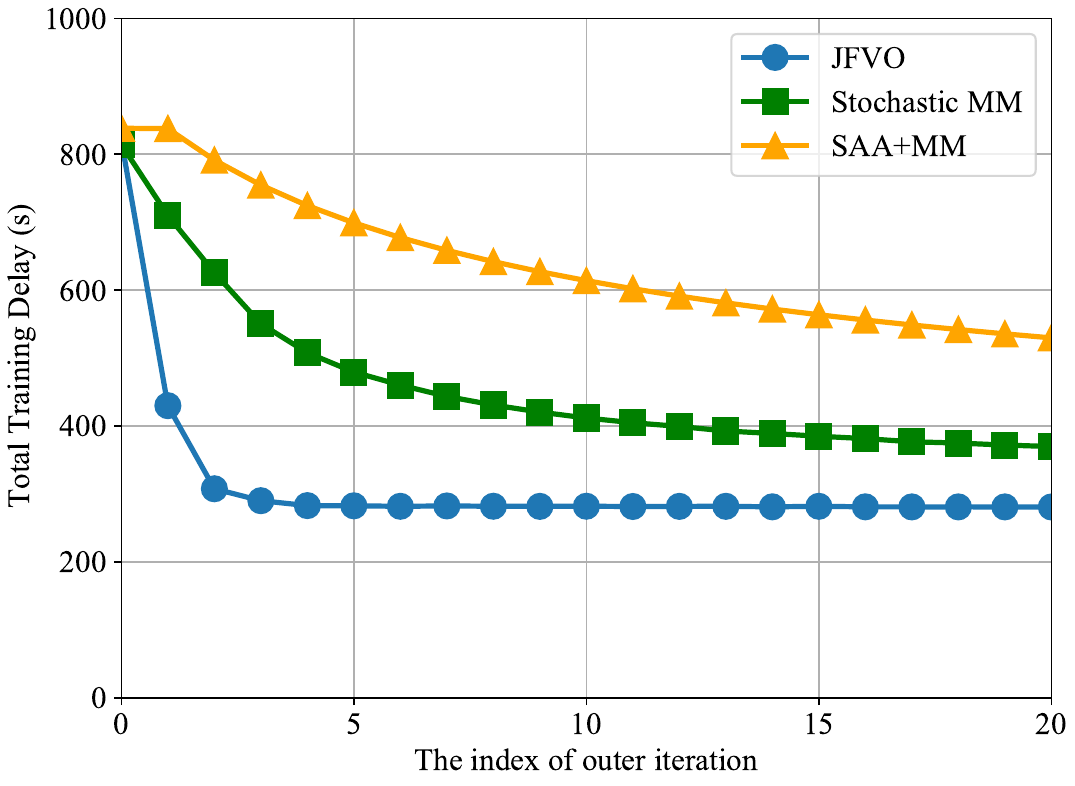}%
\label{fig_JFVO_convergence_algorithm}}
\\
\subfloat[\footnotesize{{Convergence  versus SNR. }}]
{\includegraphics[width=7.3cm,height=5.5cm]{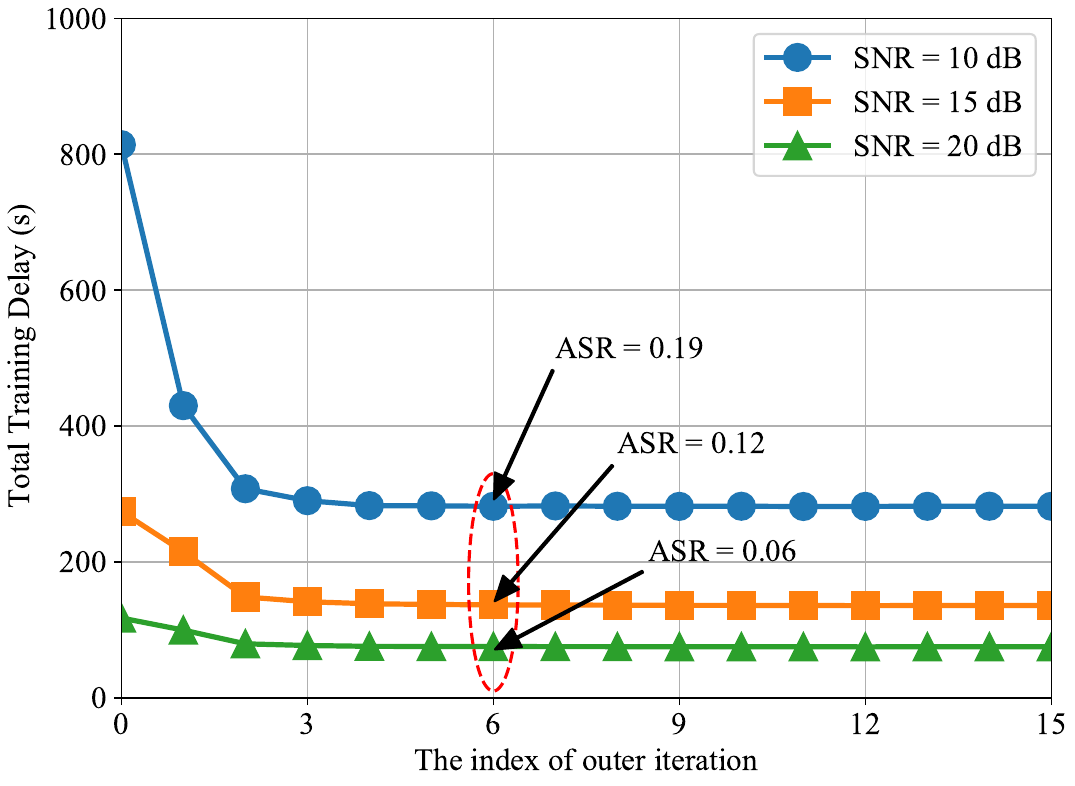}%
\label{fig_JFVO_convergence_SNR}}
\caption{{Convergence of the JFVO algorithm. Average Sharing Ratio (ASR) is the weighted average of local sharing ratio of cluster heads, which measures the degree of data leakage.  $ASR = \sum\limits_{m \in \mathcal{M}} {\frac{{\left| {{\mathcal{C}_m}} \right|}}{K} \cdot \frac{{n_m^s}}{{{n_m}}}} $. }}
\vspace{-1.5mm}
\label{fig_JFVO_convergence}
\vspace{-4mm}
\end{center}
\end{figure}

%\begin{figure}[t]
%\begin{center}
%\includegraphics[width=7.3cm,height=5.5cm]{fig-sub2.pdf}
%\vspace{-3mm}
%\caption{Convergence of the JFVO algorithm versus SNR. \hl{Average Local Sharing Ratio (ALSR) is the weighted average of local sharing ratio of cluster heads, which measures the degree of data leakage.  $ALSR = \sum\limits_{m \in \mathcal{M}} {\frac{{\left| {{\mathcal{C}_m}} \right|}}{K} \cdot \frac{{n_m^s}}{{{n_m}}}} $. } }
%\label{fig_JFVO}
%\vspace{-4mm}
%\end{center}
%\end{figure}

\emph{3)  Convergence of the JFVO Algorithm:}

{In Fig. {\protect\ref{fig_JFVO_convergence}}, we compare the convergence of the proposed JFVO algorithms with traditional approaches and verify its robustness in different  Signal-Noise-Ratio (SNR) environments. As shown in Fig. {\protect\ref{fig_JFVO_convergence}}(a), JFVO consistently outperforms baseline methods in both convergence speed and optimization results, showcasing its ability to effectively manage trade-offs between resource allocation and FL training. This superior performance can be attributed to JFVO's effectiveness in tackling stochastic non-convex objectives, making it a practical and scalable solution for heterogeneous wireless FL environments.

Furthermore, we evaluate the performance of the JFVO algorithm under SNRs of 10 dB, 15 dB, and 20 dB. As shown in  Fig. {\protect\ref{fig_JFVO_convergence}}(b), the total training delay keeps decreasing  as the iteration index increases, and ultimately converges within 6 iterations. Therefore, it substantiates the rapid convergence capability of the JFVO algorithm. Additionally, we also record the shared data volume as the algorithm converges, and the results show that the proposed framework can achieve significant performance gains of delay with little data leakage (less than 1.0).  Interestingly, as the SNR of downloading increases, the shared data volume also increases leading to more privacy leakage.  This is because that download delay at low SNR is high which requires more shared data volume to reduce the number of communication rounds.
 }

\begin{figure}[t]
\begin{center}
\includegraphics[width=7.3cm,height=5.5cm]{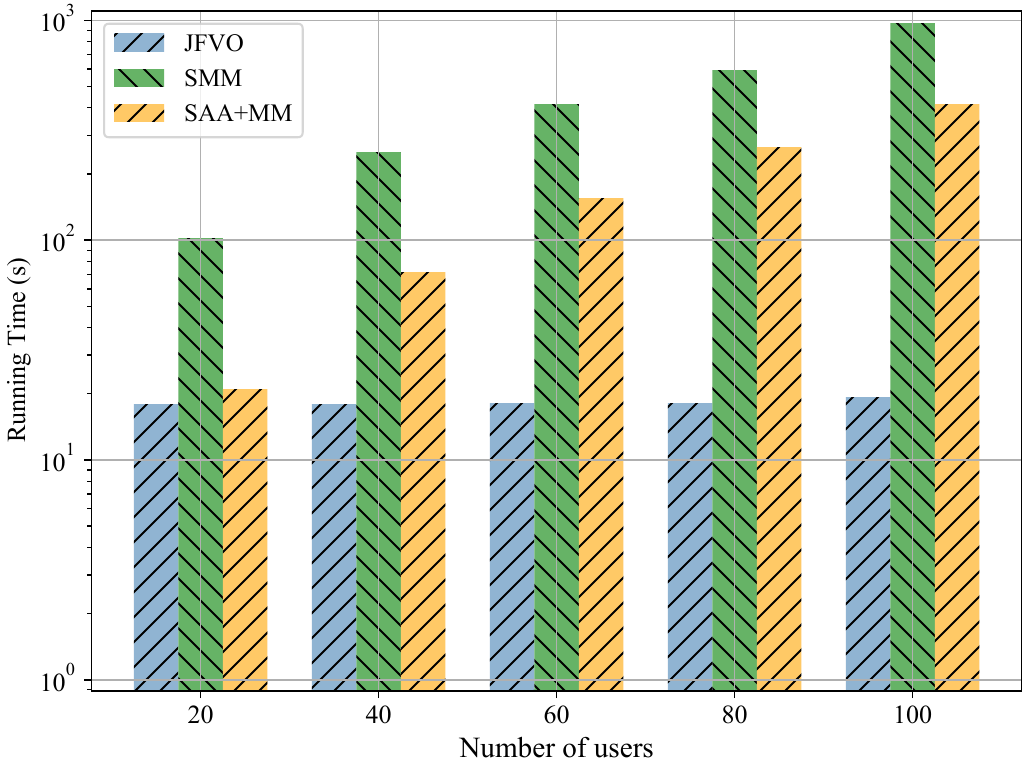}
\vspace{-3mm}
\caption{{The complex comparison of resource allocation algorithms.}}
\vspace{-3mm}
\label{fig_JFVO_RunningTime}
\end{center}
\end{figure}

{
\emph{4) Complexity Comparison of Optimization Algorithms:}

In addition to the computational complexity analysis, we evaluate the algorithm complexity is evaluated based on CPU running time, which are measured on the same machine (Intel Xeon Gold 6139M). For fairness, the start and end times of each process are recorded to compare the computational efficiency of JFVO against traditional approaches. As illustrated in Fig. {\protect\ref{fig_JFVO_RunningTime}}, JFVO demonstrates low and stable computational overhead compared to baseline algorithms. This efficiency is attributed to its recursive convex approximation function, which enables rapid convergence to optimal solutions while significantly reducing inter-iteration times. Furthermore, as the number of users increases, the running times of the SAA and SMM algorithms grow significantly, while the running time of JFVO remains nearly constant. This result shows superior capability of JFVO in handling large-scale optimization problems effectively.}

\begin{figure}[t]
\begin{center}
\subfloat[\footnotesize{{Shared data volume and SINR of each cluster}}]
{\includegraphics[width=7.3cm,height=5.5cm]{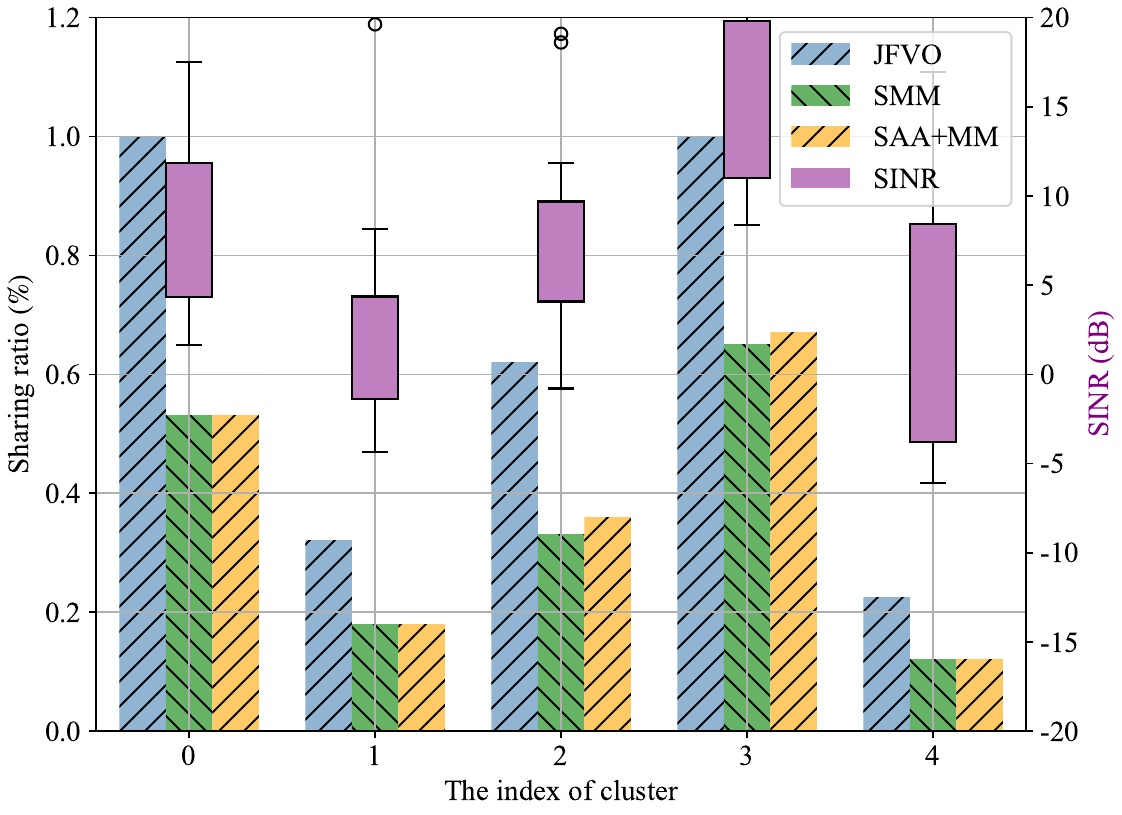}%
\label{fig_JFVO_ResourceAllocation}}
\\
\subfloat[\footnotesize{{Total delay of FL versus number of users.}}]{\includegraphics[width=7.3cm,height=5.5cm]{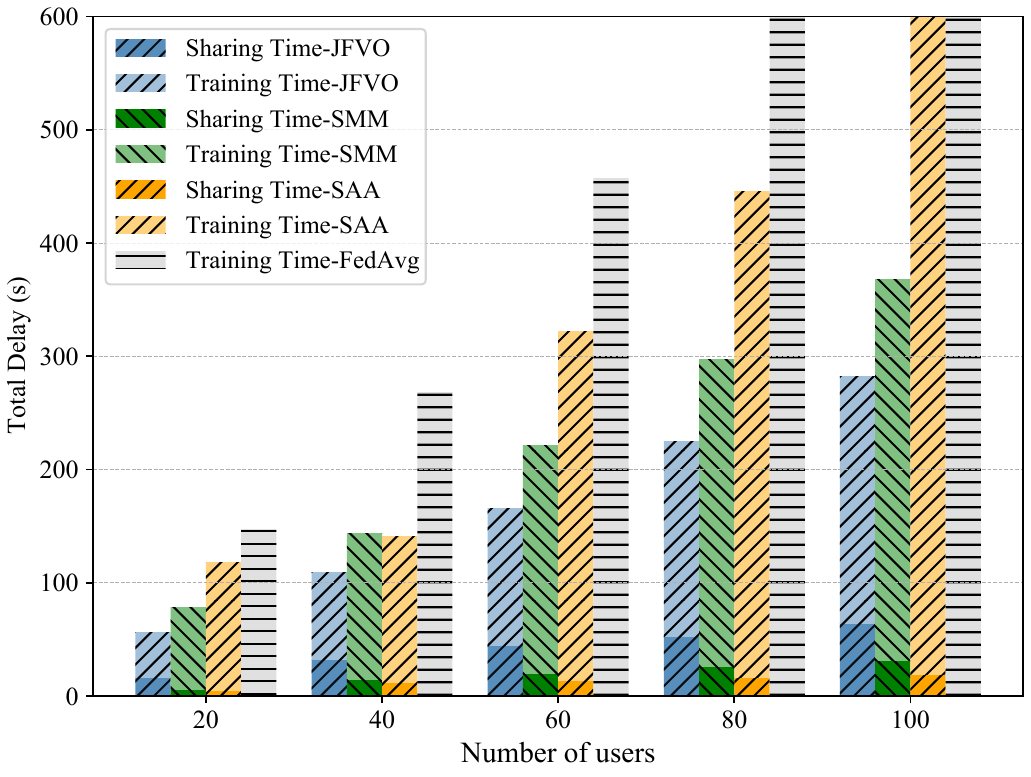}%
\label{fig_JFVO_TotalDealy}}
\caption{{Resource allocation of clustered data sharing FL}}
\vspace{-1.5mm}
\label{fig_FL_resource_analysis}
\vspace{-4mm}
\end{center}
\end{figure}

\emph{5) Resource Allocation of the Proposed Framework:}

{
We record the shared data volume for each cluster head and the SINR within the clusters. As shown in Fig. {\protect\ref{fig_FL_resource_analysis}}(a), the sharing data ratio of the cluster head decreases as the multicasting SINR diminishes. This is because the delay of data sharing is determined by the cluster with the lowest multicast rate, allowing clusters with higher SINR to exchange more data within the same multicast delay. Compared with other algorithms, JFVO can allocate more shared data in each cluster, which reduces statistical heterogeneity more effectively in FL.}

{The Fig. {\protect\ref{fig_FL_resource_analysis}}(b)  provides a detailed analysis for delay of proposed framework on communication and computation. As shown in Fig. {\protect\ref{fig_FL_resource_analysis}(b)}, the proposed JFVO algorithm outperforms both SMM and SAA in minimizing the total delay under resource-constrained conditions. This is due to its ability to effectively balance the trade-off between muticasting and FEL training latencies by jointly optimizing shared data volume and computed frequency. Notably, the shared data time of JFVO slightly increases as the number of users grows. This result is caused by increased interference among cluster heads, which leads to a lower multicast rate of data sharing. The FEL training delay (including broadcasting, computation and uploading) is less than other algorithms. This improvement stems from the reduction in the number of convergence rounds required for FEL.
}

\section{Conclusion and Future Work} \label{sec_Clusion}
%\addtolength{\topmargin}{0.111in}
\vspace{-1mm}
%In this work, we investigate and quantify the statistical heterogeneity which is identified as a primary impediment to the convergence of FL. Furthermore, we introduce a clustered data sharing framework to address the non-IID challenge in FL and formulate an optimization problem which is aimed at minimize the overall delay in FL transmission and computation. Through problem decomposition, we decompose the original problem into two subproblem, each dedicated to identifying the optimal clustering strategy and efficient resource allocation. For the clustering strategy, we establish three conditions that aid in clustering process, rooted in a privacy-preserving constrained graph. To enhance the efficacy of data sharing, a clustering algorithm is proposed for selecting cluster heads and credible associates based on the data distribution and the constraints. For resource allocation, we theoretical analyze and experimental fit the relationship between data distribution and objective function. Then due to the non-convex objective function with uncertain parameters, a stochastic optimization approach is employed to determine the optimal computed frequency and shared data volume. The experimental results show that our method performs well not only in terms of accuracy with resource-constrained environments but also in different non-IID cases.
In this work, we introduce a clustered data sharing framework to address the non-IID challenge in FEL and formulate an optimization problem which is aimed at minimize the overall delay in transmission and computation. We decompose the original problem into two subproblem, each dedicated to identifying the optimal clustering strategy and efficient resource allocation. Three conditions are established to aid in clustering process, rooted in a privacy-preserving constrained graph. To enhance the efficacy of data sharing, a clustering algorithm is proposed for selecting cluster heads and credible associates based on the data distribution and the constraints. We derive a upper bound on FEL convergence to fit the relationship between data distribution and the communication rounds. Then a stochastic optimization approach is employed to determine the optimal computed frequency and shared data volume. The experimental results show that our method exhibits well performance not only in  resource-constrained environments but also in different non-IID cases.

To further enhance model accuracy, we propose the following two potential directions: i) Efficiently share high-quality synthetic data: Sharing more high-quality data can improve accuracy further, However, this approach may trade off communication efficiency for enhanced model performance. ii) Integrate model-based approaches: Combining our framework with techniques like FedProx or SCAFFOLD can further address heterogeneity, complementing our approach.
\section*{Acknowledgment}

This work was supported in part by the National Key R\&D Program of China (No. 2021YFB3300100), and the National Natural Science Foundation of China (No. 62171062) and (U24A20234). 

\section*{Appendix A}\label{proof_1}
\emph{To prove the contrapositive of Theorem \ref{thm}:}  Assume that ${\cal M}'$ and ${\cal C}'$ (${\cal M}' \ne {{\cal M}^*}$, ${\cal C}' \ne {{\cal C}^*}$) not satisfy condition 2, i.e., there exists a cluster member ${c_0} \in {{\cal C}'_{{m_0}}}$ whose ${D_{{\rm{EMD}}}}\left( {{c_0}} \right)$ is lower than its cluster heads ${m_0} \in {\cal M}'$. Then we can set ${c_0}$ as the cluster head for a new extra cluster, making the ${\tilde D_{{\rm{EMD}}}}$ lower. Therefore, ${\cal M}'$ and ${\cal C}'$ are not the optimal solutions of the problem $\mathcal{P}$. Since ${\cal M}'$ and ${\cal C}'$ satisfy condition 2 but not satisfy condition 3, the data distributions of the cluster heads are all relatively close to the global distribution. Therefore, the greater the difference in distribution distance between nodes and cluster heads, the greater the shared gain that can be obtained, making ${\tilde D_{{\rm{EMD}}}}\left( {{\cal M}',{\cal C}'} \right) > {\tilde D_{{\rm{EMD}}}}\left( {{{\cal M}^*},{{\cal C}^*}} \right)$. Hence, ${\cal M}'$ and ${\cal C}'$ are not the optimal solutions to the problem $\mathcal{P}1'$.

\section*{Appendix B}
\subsection{Assumptions and Preliminaries}
\begin{assumption}(\emph{L-smooth}) \label{Assumption1}
For all ${w_1}$ and ${w_2}$, ${F_k}\left( {{w_1}} \right) \leqslant {F_k}\left( {{w_2}} \right) + {\left( {{w_1} - {w_2}} \right)^T}\nabla {F_k}\left( {{w_2}} \right) + \frac{L}{2}\left\| {{w_1} - {w_2}} \right\|_2^2$.
\end{assumption}

\begin{assumption}(\emph{Convex}) \label{Assumption2}
For all ${w_1}$ and ${w_2}$, ${F_k}\left( {{w_1}} \right) \geqslant {F_k}\left( {{w_2}} \right) + {\left( {{w_1} - {w_2}} \right)^T}\nabla {F_k}\left( {{w_2}} \right)$.
\end{assumption}

\begin{assumption}(\emph{Unbiasedness and Bounded Variance of local stochastic gradient}) \label{Assumption3}\

${\mathbb{E}_{\left( {x,y} \right) \sim {\mathcal{D}_k}}}\left[ {\nabla {F_k}\left( {\left( {x,y} \right),w} \right)} \right] = \nabla {F_k}\left( w \right)$,

$\mathbb{E}\left[ {{{\left\| {\nabla {F_k}\left( {\left( {x,y} \right),{w_k}} \right) - \nabla {F_k}\left( {{w_k}} \right)} \right\|}^2}} \right] \leqslant {\sigma ^2}$.
\end{assumption}

\begin{assumption}(\emph{Uniformly Bounded of local stochastic gradients}) \label{Assumption4}
$\left\| {\nabla {F_k}\left( {\left( {x,y} \right),w} \right)} \right\| \leqslant G$ for all $k = 1,...,K$.
\end{assumption}

At any round $t$, each client $k$ takes $E$ local SGD steps with learning rate $\eta$, i.e.,
$w_k^{\left( {t,i + 1} \right)} = w_k^{\left( {t,i} \right)} + \eta \nabla {F_k}\big( {\left( {x,y} \right),w_k^{\left( {t,i} \right)}} \big)$, $\forall i = 1,...,E$, $\forall k = 1,..,K$.
For the sake of simplicity, we set the same ${n_k}$ for all clients and the personal heterogeneity as ${D_k} = \sum\nolimits_y {\left\| {{P_k}\left( y \right) - {P_g}\left( y \right)} \right\|} $. Then like the works \cite{Convergence_of_FedAvg_noniid,Survey_Field_Guide_FL}, we also use the shadow sequence for clearly analyzing: the virtual global model for each step update ${\bar w^{\left( {t,i} \right)}}$, and the average EMD $\bar D$ can be described as:
\begin{equation*}
  {{\bar w^{\left( {t,i} \right)}} = \frac{1}{K}\sum\limits_k {w_k^{\left( {t,i} \right)}} ,\bar D = \frac{1}{K}\sum\limits_k {{D_k}}. }
\end{equation*}
Given this notation, we have
\begin{equation*}
  {{\bar w^{\left( {t,i + 1} \right)}} = {\bar w^{\left( {t,i} \right)}} - \frac{\eta }{K}\sum\limits_{k = 1}^K {\nabla {F_k}\left( {\left( {x,y} \right),w_k^{\left( {t,i} \right)}} \right)}.}
\end{equation*}
It is noteworthy that at the end of each round, we have ${\bar w^{(t,E)}} = {\bar w^{\left( {t + 1,0} \right)}} = w_k^{(t + 1,0)}$.

\subsection{Key Lemmas and Theorem}
\textbf{PROOF SKETCH:} First we prove two lemmas, the Lemma \ref{lemma1} gives an upper bound on the convergence of one training round, which shows that the training error of FL comes from two parts, stochastic gradient descent and model drift. Then lemma \ref{lemma2} gives an upper bound on the error of model drift, which is mainly affected by two main factors, the number of local training rounds and the local data distribution (data heterogeneity). Finally we use these two lemmas to derive an upper bound on the FL convergence of the $T$-rounds.

\begin{lemma}[\emph{Results of one round}]\label{lemma1}Assuming the client learning rate satisfies $\eta  \leqslant \frac{1}{{4L}}$, then\

{\small{
$\begin{gathered}
  \mathbb{E}\left[ {\frac{1}{E}\sum\limits_{i = 1}^E {F\left( {{{\bar w}^{\left( {t,i} \right)}}} \right) - F\left( {{w^*}} \right)} } \right] \hfill \\
   \leqslant \frac{1}{{2\eta E}}\left( {{{\left\| {{{\bar w}^{\left( {t,0} \right)}} - {w^*}} \right\|}^2} - \mathbb{E}\left[ {{{\left\| {{{\bar w}^{\left( {t,E} \right)}} - {w^*}} \right\|}^2}} \right]} \right) \hfill \\
   + \underbrace {\frac{{\eta {\sigma ^2}}}{K} + \frac{L}{{KE}}\sum\limits_{k = 1}^K {\sum\limits_{i = 1}^E {\mathbb{E}\left[ {{{\left\| {w_k^{\left( {t,i} \right)} - {{\bar w}^{\left( {t,i} \right)}}} \right\|}^2}} \right]} } .}_{\textcolor{red}{\text{model drift}}} \hfill \\
\end{gathered} $ }}
\end{lemma}

\begin{remark}
Lemma \ref{lemma1} analyzes the convergence bound of a FL round from two terms: the convergence error of stochastic gradient optimization, and the drift between the local model and the virtual global model.
\end{remark}

\begin{lemma}[\emph{Bounded model drift}]\label{lemma2}Assuming the client learning rate satisfies $\eta  \leqslant \frac{1}{{4L}}$, then\

$\begin{gathered}
  \mathbb{E}\left[ {{{\left\| {w_k^{\left( {t,i} \right)} - {{\bar w}^{\left( {t,i} \right)}}} \right\|}^2}} \right] \hfill \\
   \leqslant 2{E^2}{\eta ^2}D_k^2{G^2} + 6E{\eta ^2}D_k^2 + 4E{\eta ^2}{\sigma^2}. \hfill \\
\end{gathered} $
\end{lemma}
\begin{remark}
Lemma \ref{lemma2} highlights that model drift is influenced by two primary factors: the local step size $E$ and the heterogeneity degree of the local dataset $D_k$. The model drift error increases with the larger degree of data heterogeneity.  An interesting observation is that model drift error escalates with the growth of the local step size. This phenomenon can be readily understood as increasing the local step size shifts the local models towards the local optimums.
\end{remark}
Combine Lemmas \ref{lemma1} and \ref{lemma2} and telescope $t$ from $0$ to $T-1$ to obtain the main theorem as follows:

\begin{theorem}(\emph{Convergence rate}) \label{theorem2}Under the aforementioned assumptions (\ref{Assumption1}-\ref{Assumption4}), if the client learning rate satisfies $\eta  \leqslant \frac{1}{{4L}}$, then\

$\begin{gathered}
  \mathbb{E}\left[ {\frac{1}{{ET}}\sum\limits_{t = 0}^{T - 1} {\sum\limits_{i = 1}^E {F\left( {{{\bar w}^{\left( {t,i} \right)}}} \right) - F\left( {{w^*}} \right)} } } \right] \hfill \\
   \leqslant \frac{A}{{2\eta ET}} + \frac{{\eta {\sigma ^2}}}{K} + 2KL{E^2}{\eta ^2}{{\bar D}^2}{G^2} \hfill \\
   + 6KLE{\eta ^2}{{\bar D}^2} + 4LE{\eta ^2}{\sigma ^2},
\end{gathered} $
\end{theorem}

where $A = \left\| {{w^{\left( {0,0} \right)}} - {w^*}} \right\|$.

\emph{proof:} By combining Lemmas \ref{lemma1}, and \ref{lemma2}, we have

{\small{
$\begin{gathered}
  {\mathbb{E}}\left[ {\frac{1}{{ET}}\sum\limits_{t = 0}^{T - 1} {\sum\limits_{i = 1}^E {F\left( {{{\bar w}^{\left( {t,i} \right)}}} \right) - F\left( {{w^*}} \right)} } } \right]
\end{gathered} $}}

{\small{
$\begin{gathered}
\leqslant \frac{1}{{2\eta ET}}\sum\limits_{t = 0}^{T - 1} {\left( {{{\left\| {{{\bar w}^{\left( {t,0} \right)}} - {w^*}} \right\|}^2} - {\mathbb{E}}\left[ {{{\left\| {{{\bar w}^{\left( {t,E} \right)}} - {w^*}} \right\|}^2}} \right]} \right)}  \hfill \\
\end{gathered} $}}

{\small{
$\begin{gathered}
   + \frac{{\eta {\sigma ^2}}}{K} + \frac{L}{K}\sum\limits_{k = 1}^K {\left( {2{E^2}{\eta ^2}D_k^2{G^2} + 6E{\eta ^2}D_k^2 + 4E{\eta ^2}{\sigma^2}} \right)}
\end{gathered} $}}

{\small{
$\begin{gathered}
   \leqslant \frac{1}{{2\eta ET}}{\left\| {{{\bar w}^{\left( {0,0} \right)}} - {w^*}} \right\|^2} + \frac{{\eta {\sigma ^2}}}{K} + 4LE{\eta ^2}{\sigma^2}
\end{gathered} $}}

{\small{
$\begin{gathered}
   + LK\left( {2{E^2}{\eta ^2}{{\left( {\sum {\frac{{{D_k}}}{K}} } \right)}^2}{G^2} + 6E{\eta ^2}{{\left( {\sum {\frac{{{D_k}}}{K}} } \right)}^2}} \right)
\end{gathered} $}}

{\small{
$\begin{gathered}
   \leqslant \frac{A}{{2\eta ET}} + \frac{{\eta {\sigma ^2}}}{K} + 2LK{E^2}{\eta ^2}{{\bar D}^2}{G^2}
\end{gathered} $}}

{\small{
$\begin{gathered}
   + 6LKE{\eta ^2}{{\bar D}^2} + 4LE{\eta ^2}{\sigma^2}.
\end{gathered} $}}

\begin{remark} \label{remark_theo2}
Theorem \ref{theorem2} shows that to attain a fixed precision $\theta$, the number of communications is $\small{T_{\theta} \!=\! {\rm O}\left( {\frac{1}{{\beta  - {D^2}}}} \right)}$, Here $\beta$ is a constant for normalization. We have another insight that the convergence rate is not continue increase as the number of users $K$ increase. Although increasing $K$ can reduce the variance due to the stochastic gradients, it also increase the error from the model drift (Lemma \ref{lemma2}).
\end{remark}

\subsection{The proof of the Lemmas}
\emph{Proof} of Lemma \ref{lemma1}:

{\small{
$\begin{gathered}
   F\left( {{{\bar w}^{\left( {t,i} \right)}}} \right) - F\left( {{w^*}} \right) = \frac{1}{K}\sum\limits_{k = 1}^K {\left( {{F_k}\left( {{{\bar w}^{\left( {t,i} \right)}}} \right) - F\left( {{w^*}} \right)} \right)}
\end{gathered} $}}

{\small{
$\begin{gathered}
   = \frac{1}{K}\sum\limits_{k = 1}^K {\left( {{F_k}\left( {{{\bar w}^{\left( {t,i} \right)}}} \right) - {F_k}\left( {w_k^{\left( {t,i - 1} \right)}} \right)} \right.}
\end{gathered} $}}

{\small{
$\begin{gathered}
  \left. { + {F_k}\left( {w_k^{\left( {t,i - 1} \right)}} \right) - F\left( {{w^*}} \right)} \right)
\end{gathered} $}}

{\small{
$\begin{gathered}
  \mathop  \leqslant \limits^a \frac{1}{K}\sum\limits_{k = 1}^K {\left[ {\left\langle {\nabla {F_k}\left( {w_k^{\left( {t,i - 1} \right)}} \right),{{\bar w}^{\left( {t,i} \right)}} - w_k^{\left( {t.i - 1} \right)}} \right\rangle } \right.}
\end{gathered} $}}

{\small{
$\begin{gathered}
   + \frac{L}{2}{\left\| {{{\bar w}^{\left( {t,i} \right)}}\! -\! w_k^{\left( {t.i - 1} \right)}} \right\|^2}\left. {\! -\! \left(\! {{w^*}\! -\! w_k^{\left(\! {t.i \! - \!1} \right)}}\! \right)\nabla {F_k}\left( {w_k^{\left( {t,i\! -\! 1} \right)}} \right)} \right]
\end{gathered} $}}

{\small{
$\begin{gathered}
   = \frac{1}{K}\sum\limits_{k = 1}^K {\left[ {\left\langle {\nabla {F_k}\left( {w_k^{\left( {t,i - 1} \right)}} \right),{{\bar w}^{\left( {t,i} \right)}} - {w^*}} \right\rangle } \right.}
\end{gathered} $}}

{\small{
$\begin{gathered}
  \left. { + \frac{L}{2}{{\left\| {{{\bar w}^{\left( {t,i} \right)}} - w_k^{\left( {t.i - 1} \right)}} \right\|}^2}} \right]
\end{gathered} $}}

{\small{
$\begin{gathered}
  \mathop  \leqslant \limits^b \frac{1}{K}\sum\limits_{k = 1}^K {\left\langle {\nabla {F_k}\left( {w_k^{\left( {t,i - 1} \right)}} \right),{{\bar w}^{\left( {t,i} \right)}} - {w^*}} \right\rangle }
\end{gathered} $}}
\begin{equation}\label{Eq_one_round}
\small{
\begin{gathered}
   + L{\left\| {{{\bar w}^{\left( {t,i} \right)}} \!-\! {{\bar w}^{\left( {t,i - 1} \right)}}} \right\|^2} + \frac{L}{K}\sum\limits_{k = 1}^K {{{\left\| {{{\bar w}^{\left( {t,i-1} \right)}}\! -\! w_k^{\left( {t.i - 1} \right)}} \right\|}^2}},\;\;
\end{gathered}
}
\end{equation}
where the former part of (a) is obtained from Assumption \ref{Assumption1} and the latter part is obtained from Assumption \ref{Assumption2},  and (b) results from the equation ${\left\| {A \! + \! B} \right\|^2} \! \leqslant \! 2{\left\| A \right\|^2} \! + \! 2{\left\| B \right\|^2}$, where $A = {\bar w^{\left( {t,i} \right)}} - {\bar w^{\left( {t,i - 1} \right)}}$ and $B = {\bar w^{\left( {t,i - 1} \right)}} - \bar w_k^{\left( {t,i - 1} \right)}$.

{\small{
$\begin{gathered}
  \mathbb{E}\left[ \frac{1}{K}\sum_{k=1}^K \langle \nabla F_k(w_k^{(t,i-1)}), \bar{w}^{(t,i)} - w^* \rangle \right]
  \end{gathered} $}}

{\small{
$\begin{gathered}
  = \mathbb{E}\left[ \frac{1}{K} \! \sum_{k=1}^K \langle \! \nabla \! F_k(w_k^{(t,\!i\!-\!1\!)}) \! - \! \nabla F_k((x,y),w_k^{(t,\!i\!-\!1\!)}), \bar{w}^{(t,i)} \! - \! w^* \rangle \right]
  \end{gathered} $}}

{\small{
$\begin{gathered}
  + \mathbb{E}\left[ \frac{1}{K}\sum_{k=1}^K \langle \nabla F_k((x,y),w_k^{(t,i-1)}), \bar{w}^{(t,i)} - w^* \rangle \right] \end{gathered} $}}

{\small{
$\begin{gathered}
  \mathop = \limits^{(a)} \! \mathbb{E} \! \left[ \! \frac{1}{K} \! \sum_{k=1}^K \! \langle \nabla \! F_k(w_k^{(t,i-1)}\!) \! - \! \nabla \! F_k(\!(x,y), \! w_k^{(t,\!i\!-\!1\!)}\!), \! \bar{w}^{(t,\!i)} \! - \! \bar{w}^{(t,\!i\!-\!1\!)} \! \rangle \! \right]
  \end{gathered} $}}

{\small{
$\begin{gathered}
  + \mathbb{E}\left[ \frac{1}{\eta} \langle \bar{w}^{(t,i-1)} - \bar{w}^{(t,i)}, \bar{w}^{(t,i)} - w^* \rangle \right]
  \end{gathered} $}}

{\small{
$\begin{gathered}
  \mathop \leqslant \limits^{(b)} \! \eta \mathbb{E}\left[ \left\| \frac{1}{K}\sum_{k=1}^K \! \left( \! \nabla \! F_k(w_k^{(t,\!i\!-\!1\!)}) \! -\! \nabla F_k((x,y),w_k^{(t,\!i\!-\!1\!)}) \right) \! \right\|^2 \! \right]
  \end{gathered} $}}

{\small{
$\begin{gathered}
  + \frac{1}{4\eta} \left\| \bar{w}^{(t,i)} - \bar{w}^{(t,i-1)} \right\|^2
  \end{gathered} $}}

{\small{
$\begin{gathered}
  + \frac{1}{2\eta} \! \left( \! \left\| \bar{w}^{(t,i-1)} \! - \! w^* \right\|^2 \! - \! \left\| \bar{w}^{(t,i-1)} \! - \! \bar{w}^{(t,i)} \right\|^2 \! - \! \left\| \bar{w}^{(t,i)} \! - \! w^* \right\|^2 \right)
  \end{gathered} $}}

{\small{
$\begin{gathered}
  \leqslant \frac{\eta \sigma^2}{K} - \frac{1}{4\eta} \left\| \bar{w}^{(t,i)} - \bar{w}^{(t,i-1)} \right\|^2
\end{gathered} $}}
%\vspace{-1cm}
\begin{flalign} \label{eq:mid_one_round}
  &\;\;\;+ \frac{1}{2\eta} \left( \left\| \bar{w}^{(t,i-1)} - w^* \right\|^2 - \left\| \bar{w}^{(t,i)} - w^* \right\|^2 \right),&
\end{flalign}
where the former part of (a) holds due to $\mathbb{E}\left[ {\nabla {F_k}\left( {\left( {x,y}  \right),w} \right) - \nabla {F_k}\left(\! w \! \right)} \right] \hspace{-2mm} = \hspace{-2mm}0$, and the latter part results from \emph{perfect square trinomial} $2\left\langle {a,b} \right\rangle \! = \! {\left\| {a \! + \! b} \right\|^2} \! -\! {\left\| a \right\|^2} \! - \! {\left\| b \right\|^2}$.

Recalling the (\ref{Eq_one_round}) and plugging (\ref{eq:mid_one_round}) into it, we have
{\small{
$\begin{gathered}
  \mathbb{E}\left( {F\left( {{{\bar w}^{\left( {t,i} \right)}}} \right) - F\left( {{w^*}} \right)} \right)
   \end{gathered} $}}

{\small{
$\begin{gathered}
   \leqslant \frac{{\eta {\sigma ^2}}}{K} \! - \! \left( \! {\frac{1}{{4\eta }} \! - \! L} \! \right) \! {\left\| {{{\bar w}^{\left(\! {t,i} \right)}} \! - \! {{\bar w}^{\left( \! {t,\!i \! - \! 1} \! \right)}}} \right\|^2}
   \! + \! \frac{L}{K}\sum\limits_{k = 1}^K {{{\! \left\| {{{\bar w}^{\left( {t,i\!-\!1} \right)}} - w_k^{\left( \! {t,\!i \! -\! 1}\! \right)}} \right\|}^2}}
   \end{gathered} $}}

{\small{
$\begin{gathered}
   + \frac{1}{{2\eta }}\left( {{{\left\| {{{\bar w}^{\left( {t,i - 1} \right)}} - {w^*}} \right\|}^2} - {{\left\| {{{\bar w}^{\left( {t,i} \right)}} - {w^*}} \right\|}^2}} \right)
   \end{gathered} $}}

{\small{
$\begin{gathered}
   \mathop \leqslant \limits^{(a)} \frac{{\eta {\sigma ^2}}}{K} + \frac{L}{K}\sum\limits_{k = 1}^K {{{\left\| {{{\bar w}^{\left( {t,i-1} \right)}} - w_k^{\left( {t.i - 1} \right)}} \right\|}^2}}
\end{gathered} $}}
\begin{flalign}
\,\,\,\, &+ \frac{1}{2\eta }\left( \left\| {\bar w^{(t,i - 1)} - w^*} \right\|^2 - \left\| {\bar w^{(t,i)} - w^*} \right\|^2 \right), &
\end{flalign}
where the client learning rate satisfies $\eta  \leqslant \frac{1}{{4L}}$.  Telescoping $i$ from $0$ to $E$ completes the proof of Lemma \ref{lemma1}.

\emph{Proof} of Lemma \ref{lemma2}:
%\begin{align} \label{Eq_Heter_Gradient}
%   &\left\| \nabla F_k(w) - \nabla F(w) \right\| \nonumber \\
%   &= \left\| \mathbb{E}_{(x,y) \sim \mathcal{P}_k}[\nabla F(x,y,w)] - \mathbb{E}_{(x,y) \sim \mathcal{P}_g}[\nabla F(x,y,w)] \right\|  \nonumber \\
%   &= \left\| \scalebox{1}{$\iint$} P_k(x,y) \nabla F(x,y,w) \, dxdy \right. \nonumber \\
%   &\quad - \left. \scalebox{1}{$\iint$} P_g(x,y) \nabla F(x,y,w) \, dxdy \right\| \nonumber \\
%   &\leqslant \scalebox{1}{$\iint$} \left\| (P_k(x,y) - P_g(x,y)) \nabla F(x,y,w) \right\| \, dxdy \nonumber \\
%   &\leqslant \scalebox{1}{$\int$}_y \left\| P_k(y) - P_g(y) \right\| \nonumber \\
%   &\quad \scalebox{1}{$\int$}_x \left\| P_k(x|y) \nabla F(x,y,w) \right\| \, dxdy \nonumber \\
%   &= \sum_{y = 1}^Y \left\| P_k(y) - P_g(y) \right\| \mathbb{E}_{x|y} \left[ \left\| \nabla F(x,y,w) \right\| \right] \nonumber \\
%   &\leqslant D_k G.
%\end{align}

$\begin{gathered}
   \left\| {\nabla {F_k}\left( w \right) - \nabla F\left( w \right)} \right\|
   \end{gathered} $

$\begin{gathered}
   = \left\| {{\mathbb{E}_{\left( {x,y} \right) \sim {{\mathcal{P}}_k}}}\left[ {\nabla F\left( {x,y,w} \right)} \right]} \right. \hfill \\
  \left. { - {\mathbb{E}_{\left( {x,y} \right) \sim {{\mathcal{P}}_g}}}\left[ {\nabla F\left( {x,y,w} \right)} \right]} \right\| \end{gathered} $

$\begin{gathered}
   = \left\| {\scalebox{1}{$\iint$} {{P_k}\left( {x,y} \right)\nabla F\left( {x,y,w} \right)}dxdy} \right.
   \end{gathered} $

$\begin{gathered}
  \left. { - \scalebox{1}{$\iint$} {{P_g}\left( {x,y} \right)\nabla F\left( {x,y,w} \right)}dxdy} \right\|
  \end{gathered} $

$\begin{gathered}
   \leqslant \scalebox{1}{$\iint$} {\left\| {\left( {{P_k}\left( {x,y} \right) - {P_g}\left( {x,y} \right)} \right)\nabla F\left( {x,y,w} \right)} \right\|}dxdy
   \end{gathered} $

$\begin{gathered}
   \leqslant \scalebox{1}{$\int_y$} {\left\| {{P_k}\left( y \right) - {P_g}\left( y \right)} \right\|}
   \end{gathered} $

$\begin{gathered}
  \scalebox{1}{$\int_x$} {\left\| {{P_k}\left( {\left. x \right|y} \right)\nabla F\left( {x,y,w} \right)} \right\|} dxdy
  \end{gathered} $

$\begin{gathered}
   = \scalebox{1}{$\int_y$} {\left\| {{P_k}\left( y \right) - {P_g}\left( y \right)} \right\|}
   \end{gathered} $

$\begin{gathered}
  \left( \scalebox{1}{$\int_x$} {\left\| {{P_k}\left( {\left. x \right|y} \right)\nabla F\left( {x,y,w} \right)} \right\|} dx \right) dy
  \end{gathered} $

$\begin{gathered}
   = \sum\limits_{y = 1}^Y {\left\| {{P_k}\left( y \right) - {P_g}\left( y \right)} \right\|{\mathbb{E}_{\left. x \right|y}}\left[ {\left\| {\nabla F\left( {x,y,w} \right)} \right\|} \right]}
   \end{gathered} $

\begin{flalign}\label{Eq_Heter_Gradient}
    & \,\,\,\, \leqslant {D_k}G. &
\end{flalign}

This inequality can be applied to other non-iid cases as well, simply by replacing the distribution function, i.e., ${P_k}\left( y \right)$.

%\begin{equation} \label{Eq_Object_Lemma2}
%{
%\small{
%\begin{gathered}
%  {\mathbb{E}}\left[ {{{\left\| {w_1^{\left( {t,i + 1} \right)} - w_2^{\left( {t,i + 1} \right)}} \right\|}^2}} \right] \hfill \\
%   = {\mathbb{E}}\left[ {\left\| {w_1^{\left( {t,i} \right)} - w_2^{\left( {t,i} \right)}} \right.} \right. \hfill \\
%  \left. { - {{\left. {\eta \left( {\nabla {F_1}\left( {\left( {x,y} \right),w_1^{\left( {t,i} \right)}} \right) - \nabla {F_2}\left( {\left( {x,y} \right),w_2^{\left( {t,i} \right)}} \right)} \right)} \right\|}^2}} \right] \hfill \\
%   \leqslant {\left\| {w_1^{\left( {t,i} \right)} - w_2^{\left( {t,i} \right)}} \right\|^2} \hfill \\
%   - 2\eta \underbrace {\left\langle {\nabla {F_1}\left( {w_1^{\left( {t,i} \right)}} \right)} \right. - \nabla {F_2}\left( {w_2^{\left( {t,i} \right)}} \right),\left. {w_1^{\left( {t,i} \right)} - w_2^{\left( {t,i} \right)}} \right\rangle }_{\textcolor{red}{\text{B}}} \hfill \\
%   + {\eta ^2}{\underbrace {{\left\| {\nabla {F_1}\left( {w_1^{\left( {t,i} \right)}} \right) - \nabla {F_2}\left( {w_2^{\left( {t,i} \right)}} \right)} \right\|}^2}_{\textcolor{red}{\text{C}}}} + 2{\eta ^2}{G^2} \hfill \\
%\end{gathered}  } }
%\end{equation}
{\small{
$\begin{gathered}
   {\mathbb{E}}\left[ {{{\left\| {w_1^{\left( {t,i + 1} \right)} - w_2^{\left( {t,i + 1} \right)}} \right\|}^2}} \right]
\end{gathered} $}}

{\small{
$\begin{gathered}
   = {\mathbb{E}}\left[ {\left\| {w_1^{\left( {t,i} \right)} - w_2^{\left( {t,i} \right)}} \right.} \right.
\end{gathered} $}}

{\small{
$\begin{gathered}
   \left. { - {{\left. {\eta \left( {\nabla {F_1}\left( {\left( {x,y} \right),w_1^{\left( {t,i} \right)}} \right) - \nabla {F_2}\left( {\left( {x,y} \right),w_2^{\left( {t,i} \right)}} \right)} \right)} \right\|}^2}} \right]
\end{gathered} $}}

{\small{
$\begin{gathered}
   \leqslant {\mathbb{E}}\left[ {{{\left\| {w_1^{\left( {t,i} \right)} - w_2^{\left( {t,i} \right)}} \right\|}^2}} \right.
\end{gathered} $}}

{\small{
$\begin{gathered}
   - 2\eta\! \left\langle \!{\nabla {F_1}\!\left( {\left(\! {x,y}\! \right),\!w_1^{\left(\! {t,i}\! \right)}}\! \right)\! -\! \nabla {F_2}\left( \!{\left( \!{x,y} \right),\!w_2^{\left( {t,i} \!\right)}} \right)} \!\right.,\left. {w_1^{\left( {t,i} \right)} \!-\! w_2^{\left(\! {t,i}\! \right)}} \right\rangle
\end{gathered} $}}

{\small{
$\begin{gathered}
   \left. { + {\eta ^2}{{\left\| {\nabla {F_1}\left( {\left( {x,y} \right),w_1^{\left( {t,i} \right)}} \right) - \nabla {F_2}\left( {\left( {x,y} \right),w_2^{\left( {t,i} \right)}} \right)} \right\|}^2}} \right]
\end{gathered} $}}

{\small{
$\begin{gathered}
   \left. { + {\eta ^2}{{\left\| {\nabla {F_1}\left( {\left( {x,y} \right),w_1^{\left( {t,i} \right)}} \right) - \nabla {F_2}\left( {\left( {x,y} \right),w_2^{\left( {t,i} \right)}} \right)} \right\|}^2}} \right]
\end{gathered} $}}

{\small{
$\begin{gathered}
   - 2\eta \left\langle {\nabla {F_1}\left( {w_1^{\left( {t,i} \right)}} \right)} \right. - \nabla {F_2}\left( {w_2^{\left( {t,i} \right)}} \right),\left. {w_1^{\left( {t,i} \right)} - w_2^{\left( {t,i} \right)}} \right\rangle
\end{gathered} $}}

{\small{
$\begin{gathered}
   + {\eta ^2}\left\| {\left( {\nabla {F_1}\left( {\left( {x,y} \right),w_1^{\left( {t,i} \right)}} \right) - \nabla {F_1}\left( {w_1^{\left( {t,i} \right)}} \right)} \right)} \right.
\end{gathered} $}}

{\small{
$\begin{gathered}
   - \left( {\nabla {F_2}\left( {\left( {x,y} \right),w_2^{\left( {t,i} \right)}} \right) - \nabla {F_2}\left( {w_2^{\left( {t,i} \right)}} \right)} \right)
\end{gathered} $}}

{\small{
$\begin{gathered}
   + {\left. {\left( {\nabla {F_1}\left( {w_1^{\left( {t,i} \right)}} \right) - \nabla {F_2}\left( {w_2^{\left( {t,i} \right)}} \right)} \right)} \right\|^2}
\end{gathered} $}}

{\small{
$\begin{gathered}
   \mathop  \leqslant \limits^a {\left\| {w_1^{\left( {t,i} \right)} - w_2^{\left( {t,i} \right)}} \right\|^2}
\end{gathered} $}}

{\small{
$\begin{gathered}
   - 2\eta \underbrace {\left\langle {\nabla {F_1}\left( {w_1^{\left( {t,i} \right)}} \right)} \right. - \nabla {F_2}\left( {w_2^{\left( {t,i} \right)}} \right),\left. {w_1^{\left( {t,i} \right)} - w_2^{\left( {t,i} \right)}} \right\rangle }_{\textcolor{red}{\text{B}}}
\end{gathered} $}}

\begin{equation} \label{Eq_Object_Lemma2}
{
\small{
\begin{gathered}
   + {\eta ^2}{\underbrace {{\left\| {\nabla {F_1}\left( {w_1^{\left( {t,i} \right)}} \right) - \nabla {F_2}\left( {w_2^{\left( {t,i} \right)}} \right)} \right\|}^2}_{\textcolor{red}{\text{C}}}} + 2{\eta ^2}{\sigma ^2}. \quad\quad\quad \hfill \\
\end{gathered}   } }
\end{equation}

We next focus on bounding $B$ and $C$:

{\small{
$\begin{gathered}
   - B =  - \left\langle {\nabla {F_1}\left( {w_1^{\left( {t,i} \right)}} \right)} \right. - \nabla {F_2}\left( {w_2^{\left( {t,i} \right)}} \right),\left. {w_1^{\left( {t,i} \right)} - w_2^{\left( {t,i} \right)}} \right\rangle
\end{gathered} $}}

{\small{
$\begin{gathered}
   \leqslant  - \left\langle {\nabla F\left( {w_1^{\left( {t,i} \right)}} \right)} \right. - \nabla F\left( {w_2^{\left( {t,i} \right)}} \right),\left. {w_1^{\left( {t,i} \right)} - w_2^{\left( {t,i} \right)}} \right\rangle
\end{gathered} $}}

{\small{
$\begin{gathered}
   + \left\langle {\nabla F\left( {w_1^{\left( {t,i} \right)}} \right)} \right. - \nabla {F_1}\left( {w_1^{\left( {t,i} \right)}} \right),\left. {w_1^{\left( {t,i} \right)} - w_2^{\left( {t,i} \right)}} \right\rangle
\end{gathered} $}}

{\small{
$\begin{gathered}
   + \left\langle {\nabla {F_2}\left( {w_2^{\left( {t,i} \right)}} \right)} \right. - \nabla F\left( {w_2^{\left( {t,i} \right)}} \right),\left. {w_1^{\left( {t,i} \right)} - w_2^{\left( {t,i} \right)}} \right\rangle
\end{gathered} $}}

{\small{
$\begin{gathered}
   \mathop  \leqslant \limits^a \frac{{ - 1}}{L}{\left\| {\nabla F\left( {w_1^{\left( {t,i} \right)}} \right) - \nabla F\left( {w_2^{\left( {t,i} \right)}} \right)} \right\|^2}
\end{gathered} $}}

{\small{
$\begin{gathered}
   + \left( {{D_1} + {D_2}} \right)G\left\| {w_1^{\left( {t,i} \right)} - w_2^{\left( {t,i} \right)}} \right\|
\end{gathered} $}}

{\small{
$\begin{gathered}
   \leqslant \frac{{ - 1}}{L}{\left\| {\nabla F\left( {w_1^{\left( {t,i} \right)}} \right) - \nabla F\left( {w_2^{\left( {t,i} \right)}} \right)} \right\|^2}
\end{gathered} $}}

\begin{equation} \label{Eq_B}
{
\small{
\begin{gathered}
   + \frac{1}{{2\eta E}}{\left\| {w_1^{\left( {t,i} \right)} - w_2^{\left( {t,i} \right)}} \right\|^2} + 2\eta E\left( {{D_1} + {D_2}} \right)G, \quad\quad\quad\quad \hfill \\
\end{gathered}  }}
\end{equation}
where the former term of inequality $(a)$ can be can be derived based on Assumptions \ref{Assumption1} and \ref{Assumption2}, and the second term is based on (\ref{Eq_Heter_Gradient}). Similarly, the $C$ is bounded as:

\begin{equation} \label{Eq_C}
{
\small{
\begin{gathered}
C = {\left\| {\nabla {F_1}\left( {w_1^{\left( {t,i} \right)}} \right) - \nabla {F_2}\left( {w_2^{\left( {t,i} \right)}} \right)} \right\|^2} \hfill \\
\leqslant \left\| {\left( {\nabla F\left( {w_1^{\left( {t,i} \right)}} \right) - \nabla F\left( {w_2^{\left( {t,i} \right)}} \right)} \right)} \right. + \left( {\nabla {F_1}\left( {w_1^{\left( {t,i} \right)}} \right)} \right. \hfill \\
- \left. {\nabla F\left( {w_1^{\left( {t,i} \right)}} \right)} \right){\left. { + \left( {\nabla F\left( {w_2^{\left( {t,i} \right)}} \right) - \nabla {F_2}\left( {w_2^{\left( {t,i} \right)}} \right)} \right)} \right\|^2} \hfill \\
\leqslant 3{\left\| {\nabla F\left( {w_1^{\left( {t,i} \right)}} \right) - \nabla F\left( {w_2^{\left( {t,i} \right)}} \right)} \right\|^2} + 3D_1^2 + 3D_2^2. \hfill \\
\end{gathered} }}
\end{equation}
Replacing the bounds (\ref{Eq_B}) and (\ref{Eq_C}) back to the (\ref{Eq_Object_Lemma2}), we have
\begin{equation}
{
\small{
\begin{gathered}
  {\mathbb{E}}\left[ {{{\left\| {w_1^{\left( {t,i + 1} \right)} - w_2^{\left( {t,i + 1} \right)}} \right\|}^2}} \right] \hfill \\
   \leqslant \left( {1 + \frac{1}{E}} \right){\left\| {w_1^{\left( {t,i} \right)} - w_2^{\left( {t,i} \right)}} \right\|^2} \hfill \\
   - \left( {\frac{{2\eta }}{L} - 3{\eta ^2}} \right)\left( {\nabla F\left( {w_1^{\left( {t,i} \right)}} \right) - \nabla F\left( {w_2^{\left( {t,i} \right)}} \right)} \right) \hfill \\
   + E{\eta ^2}{G^2}{\left( {{D_1} + {D_2}} \right)^2} + 3{\eta ^2}D_1^2 + 3{\eta ^2}D_2^2 + 2{\eta ^2}{\sigma ^2} \hfill \\
  \mathop  \leqslant \limits^a \left( {1 + \frac{1}{E}} \right){\left\| {w_1^{\left( {t,i} \right)} - w_2^{\left( {t,i} \right)}} \right\|^2} \hfill \\
   + E{\eta ^2}{G^2}{\left( {{D_1} + {D_2}} \right)^2} + 3{\eta ^2}D_1^2 + 3{\eta ^2}D_2^2 + 2{\eta ^2}{\sigma ^2}, \hfill \\
\end{gathered}  }}
\end{equation}
where (a) holds due to $\eta  \leqslant \frac{1}{{4L}}$. Then it can be derived by conducting a geometric series:
\begin{equation}
{
\small{
\begin{gathered}
  {\mathbb{E}}\left[ {{{\left\| {w_1^{\left( {t,i} \right)} - w_2^{\left( {t,i} \right)}} \right\|}^2}} \right] \hfill \\
   \leqslant \frac{{{{\left( {1 \!+\! \frac{1}{E}} \right)}^i}\! - \!1}}{{{1 \mathord{\left/
 {\vphantom {1 E}} \right.
 \kern-\nulldelimiterspace} E}}} \! \left[ {E{\eta ^2}{G^2}{{\left( {{D_1}\! +\! {D_2}} \right)}^2}\! +\! 3{\eta ^2}D_1^2 \!+\! 3{\eta ^2}D_2^2 \!+\! 2{\eta ^2}{\sigma^2}} \right] \hfill \\
   \leqslant 2{E^2}{\eta ^2}{G^2}{\left( {{D_1} + {D_2}} \right)^2} + 6E{\eta ^2}D_1^2 + 6E{\eta ^2}D_2^2 + 4E{\eta ^2}{\sigma^2}. \hfill \\
\end{gathered} }}
\end{equation}

Replacing $w_1^{\left( {t,i} \right)}$ as $w_k^{\left( {t,i} \right)}$, and $w_2^{\left( {t,i} \right)}$ as ${{\bar w}^{\left( {t,i} \right)}}$, we have

$\begin{gathered}
   \mathbb{E}\left[ {{{\left\| {w_k^{\left( {t,i} \right)} - {{\bar w}^{\left( {t,i} \right)}}} \right\|}^2}} \right] \hfill \\
   \leqslant 2{E^2}{\eta ^2}{G^2}D_k^2 + 6E{\eta ^2}D_k^2 + 4E{\eta ^2}{\sigma^2}, \hfill \\
\end{gathered} $

and the proof is completed.

\footnotesize
\bibliographystyle{IEEEtran}
\bibliography{IEEEexample}%参考文献文件名，不需要后缀
\end{document}